\documentclass[10pt,twocolumn,letterpaper]{article}
\usepackage{wacv}
\makeatletter
\@namedef{ver@everyshi.sty}{}
\makeatother
\usepackage{times}
\usepackage{epsfig}
\usepackage{graphicx}
\usepackage{amsmath}
\usepackage{amssymb}
\usepackage{subfigure}
\usepackage{algorithmic}
\usepackage[ruled,vlined]{algorithm2e}
\usepackage{comment}
\usepackage{enumitem}
\usepackage{appendix}
\usepackage[accsupp]{axessibility}
\setlist[enumerate]{itemsep=0mm}

\usepackage{tikz}

\newcommand\copyrighttextfinal{%
    
    \scriptsize\copyright\ 2022 IEEE. Personal use of this material is permitted. Permission from IEEE must be obtained for all other uses, in any current or future media, including reprinting/republishing this material for advertising or promotional purposes, creating new collective works, for resale or redistribution to servers or lists, or reuse of any copyrighted component of this work in other works.}%
\newcommand\copyrightnotice{%
    
    \begin{tikzpicture}[remember picture,overlay]%
        
        \node[anchor=south,yshift=10pt] at (current page.south) {{\parbox{\dimexpr\textwidth-\fboxsep-\fboxrule\relax}{\copyrighttextfinal}}};%
    \end{tikzpicture}%
    
}
\def\wacvPaperID{0041} 

\wacvfinalcopy 

\wacvapplicationstrack

\ifwacvfinal
\usepackage[breaklinks=true,bookmarks=false]{hyperref}
\else
\usepackage[pagebackref=true,breaklinks=true,colorlinks,bookmarks=false,
           colorlinks=true,   
           pdfusetitle=true]{hyperref}

\fi

\ifwacvfinal
\else
\fi

\begin{document}

\title{Knowing What to Label for Few Shot Microscopy Image Cell Segmentation}

\author{Youssef Dawoud $^{1}$\footnote{Contact Author}  \and Arij Bouazizi $^{1,2}$ \and Katharina Ernst $^{3}$ \and Gustavo Carneiro $^{4}$ \and Vasileios Belagiannis $^{5}$\\
\\
$^1$ Universit\"at Ulm, Ulm, Germany\\
$^2$ Mercedes-Benz AG, Stuttgart, Germany\\
$3$ Ulm University Medical Center, Ulm, Germany\\
$4$ Centre for Vision, Speech and Signal Processing, University of Surrey, United Kingdom\\
$5$ Friedrich-Alexander-Universität Erlangen-Nürnberg, Erlangen, Germany \\
{\tt\small youssef.dawoud@uni-ulm.de}
}

\maketitle

\begin{abstract}

 In microscopy image cell segmentation, it is common to train a deep neural network on source data, containing different types of microscopy images, and then fine-tune it using a support set comprising a few randomly selected and annotated training target images. In this paper, we argue that the random selection of unlabelled training target images to be annotated and included in the support set may not enable an effective fine-tuning process, so we propose a new approach to optimise this image selection process. Our approach involves a new scoring function to find informative unlabelled target images. In particular, we propose to measure the consistency in the model predictions on target images against specific data augmentations. However, we observe that the model trained with source datasets does not reliably evaluate consistency on target images. To alleviate this problem, we propose novel self-supervised pretext tasks to compute the scores of unlabelled target images. Finally, the top few images with the least consistency scores are added to the support set for oracle (i.e., expert) annotation and later used to fine-tune the model to the target images. In our evaluations that involve the segmentation of five different types of cell images, we demonstrate promising results on several target test sets compared to the random selection approach as well as other selection approaches, such as Shannon's entropy and Monte-Carlo dropout. 
\end{abstract} 
\section{Introduction}
\copyrightnotice
Microscopy image cell segmentation is one of the main fields in the area of medical image with the focus of studying the morphological properties of biological cells, i.e. geometrical shape and size along with other tasks such as cell detection \cite{zhang2012classifying}, segmentation \cite{ciresan2012deep} and counting \cite{arteta2016counting,dijkstra2018centroidnet}. Over the past years, research efforts have been devoted to automate microscopy image cell analysis, initially, with the support of classical image processing algorithms \cite{xing2016robust}. Subsequently, deep neural networks (DNNs), specifically, encoder-decoder architectures have dramatically evolved to become the state-of-the art automation approach in several microscopy image cell tasks, including microscopy image cell segmentation \cite{ciresan2012deep}. 

\begin{figure*}[!t]
\centering
	
	\rotatebox[origin=l]{90}{TNBC}
	\includegraphics[width=3.75cm]{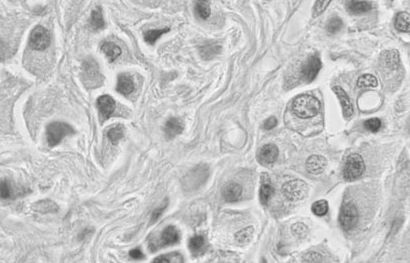}
	\includegraphics[width=3.75cm]{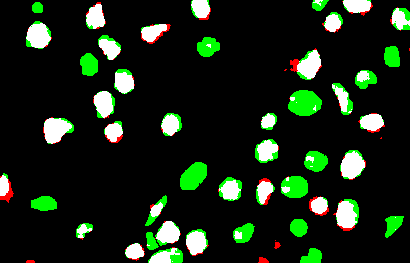}
	\includegraphics[width=3.75cm]{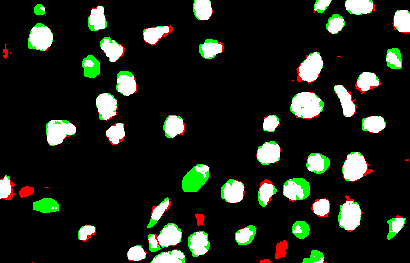}

	\rotatebox[origin=l]{90}{ssTEM}
	\includegraphics[width=3.75cm]{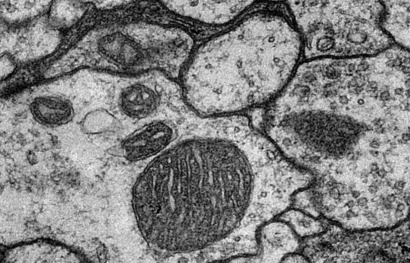}
	\includegraphics[width=3.75cm]{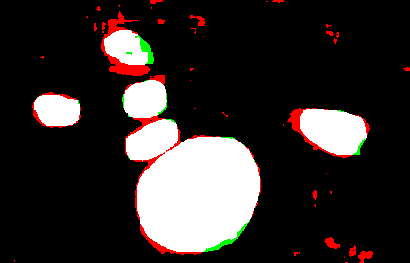}
	\includegraphics[width=3.75cm]{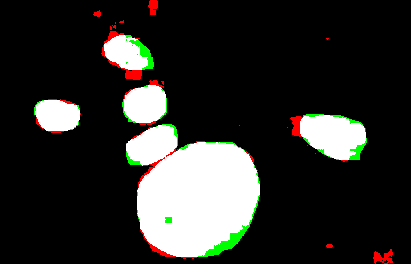}

	
	\rotatebox[origin=l]{90}{EM}
	\subfigure[c][Input]{\includegraphics[width=3.75cm]{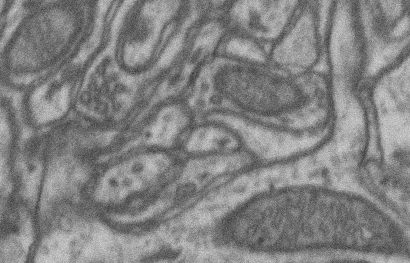}}
	\subfigure[c][Random]{\includegraphics[width=3.75cm]{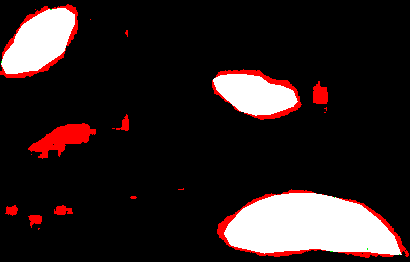}}
    \subfigure[c][Ours]{\includegraphics[width=3.75cm]{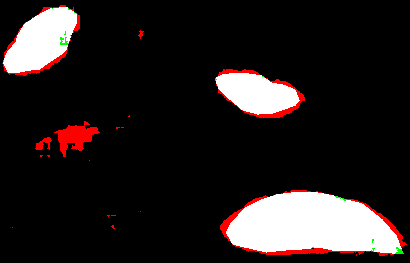}}

	\caption{Visual Result. We visually compare our scoring function (Ours) to random selection using the FCRN architecture at $|\mathcal{B}| = 3$-shots.  The red colour corresponds to false positive, the green colour to false negative, the black colour to true negative, and the white colour to true positive. Best viewed in colour.}
	\label{fig:VisualComp1}
\end{figure*}

Earlier studies employed DNNs to learn a fully supervised cell segmentation model, which required the collection and pixel-level labelling of a large amount of microscopy image data to enable a robust training \cite{xie2018microscopy}. Recently, a more practical study~\cite{Dawoud2020} showed a cell-segmentation method that can be trained with a support set containing a few annotated microscopy training images, this method is known as few-shot microscopy image cell segmentation. In this setup, a deep neural network model is trained using source data, containing training images from various types of cell segmentation problems. Afterwards, the trained model is fine-tuned to the target images with cells of interest, using a support set containing few randomly selected and annotated microscopy images. Even though effective, random image selection can be improved since the informativeness of the support set may be low, which may result in a poor fine-tuning process that also leads to a low-accuracy performance in the unseen testing target images.

In this paper, we propose a new approach to optimise the selection of samples to annotate and include in the support set, so that it contains highly informative training samples that help improve the classification accuracy of the fine-tuned model, compared with random selection. In return, we offer a more efficient use of expert's time for annotation. Our selection approach relies on a scoring function to select the support set from the unlabelled target set by evaluating the consistency in the predictions of the segmentation model. In particular, our scoring function calculates the pixel-wise cross-entropy loss between the segmentation model prediction using an image from the target set and the predictions of its augmented versions. However, we notice that the segmentation model trained using only the source data does not produce a reliable scoring function for the images belonging to the target set. Therefore, we propose to fine tune the segmentation model to the unlabelled target set using novel pretext tasks. Specifically, we propose to learn cell segmentation in images belonging to the target set using pseudo-binary segmentation labels, which we generate using classical image processing segmentation operations. The fine-tuned model, obtained from this pseudo-segmentation learning, is used to calculate the cross entropy scores of the target images. Then, the top few images with the least consistency scores are added to the support set for oracle (i.e., expert) annotation and later used to fine-tune the model to the target images. At last, we evaluate the fine-tuned model on the testing target images. We summarise our contribution as follows 1) We propose a novel pretext task of cell segmentation learning to fine-tune the segmentation model which we use afterwards for selecting samples to be annotated in a few-shot learning problem 2) We present a new scoring function for support set selection from the unlabelled target data that measures the performance consistency for the pretext tasks as a function of specific data augmentations 3) In our experiments we show promising results on five target sets involving different types of cell images compared to the random selection approach, in addition to other selection approaches, namely, Shannon's entropy and Monte-Carlo dropout. Our code and models are made publicly available. \footnote{ \url{https://github.com/Yussef93/KnowWhatToLabel/}}

\section{Related Work}

\subsection{Cell Segmentation}
Early automatic microscopy cell segmentation methods have been developed with the aid of classical image processing and computer vision algorithms~\cite{faustino2009automatic,lu2015improved,wahlby2004combining}. More recently, deep neural network architectures ranging from fully convolutional networks (FCN) to self-attention based have significantly evolved to become the state-of-the-art automation approach for several cell analysis tasks like nuclei segmentation \cite{naylor2017nuclei}, mitosis detection in histology images \cite{cirecsan2013mitosis} and cardiac segmentation in MRI images \cite{gao2021utnet}. Among FCN architectures, U-Net~\cite{ronneberger2015u}, was firstly developed for the task of neuronal structure segmentation in electron microscopy, but nowadays, it is applied in several types of medical image analysis problems. Another type of FCN architecture is the fully convolutional regression network (FCRN)~\cite{xie2018microscopy}, designed for cell counting and segmentation in microscopy images. In our work, we rely on FCRN architecture for automating microscopy cell segmentation.

\subsection{Few-Shot Segmentation}

The availability of large annotated training sets enables a robust fully supervised training of models, but many real-world problems only contain small training sets, reducing the viability of supervised-learning. These problems are known as few-shot learning. To that end, several approaches have been developed to enable the learning of a generic model that can be adapted to different tasks using a limited amount of annotated training data \cite{finn2017model,nichol2018}. Only few studies exist on the problem of few-shot medical image segmentation \cite{mondal2018few,ouyang2020self}. Among them is few-shot microscopy image cell segmentation \cite{Dawoud2020} and organ segmentation\cite{makarevich2021metamedseg}, where a model is trained using source datasets and then fine-tuned to the target images using a support set containing a handful of randomly selected and annotated images. Nevertheless, the randomly selected images in the support set may not be informative for the fine-tuning process, which can result in poor training and low segmentation accuracy in the target microscopy testing images. Therefore, we argue that the selection of images to be included in the support set must be optimised, in terms of their information content, to achieve a better fine-tuning process that can result in a good cell segmentation performance in target microscopy testing images.

\subsection{Self-Supervised Learning in Medical Image Analysis}

Learning prior representations from the unlabelled data, i.e. self-supervision, has proven to be an effective approach when fine-tuned on subsequent target tasks such as classification and segmentation. Numerous approaches have been developed for learning prior representations in deep neural networks, e.g., contrastive learning \cite{chen2020simple}, jigSaw puzzles \cite{noroozi2016unsupervised}, rotation prediction \cite{gidaris2018unsupervised} and image reconstruction \cite{baldi2012autoencoders}. A wide variety of generic self-supervised tasks have been adapted to various medical image analysis applications, like jigSaw and rotation prediction for 3D computed tomography (CT) scans \cite{taleb20203d}, contrastive learning for spinal MRI \cite{jamaludin2017self} and context restoration \cite{chen2019self} for CT and ultrasound scan analysis. However, other studies have proposed target specific pretext tasks \cite{dawoud2022edge}, for instance, superpixels prediction for abdominal CT-scans \cite{ouyang2020self}. In our work, we follow a similar path as in \cite{ouyang2020self}, where we propose a new target specific pretext task. More specifically, we use conventional image segmentation operations to extract binary segmentation pseudo-labels for the microscopy images in the target dataset. Afterwards, we fine-tune our generic segmentation model using the pseudo-labelled target dataset to obtain a good prior representation for the target data, which we leverage to design our scoring function.

\subsection{Support Set Selection Approaches}
The selection of data samples has been thoroughly studied under different learning paradigms, but more noticeably in active learning \cite{settles2009active}. Most studies rely on scoring functions for selecting images to be annotated from a pool of unlabelled images, where data samples fulfilling the selection criterion are annotated and appended to the labelled set under the constraint of a limited annotation budget. Then, deep neural networks are trained on a target learning task at hand, e.g. classification or segmentation, using this labelled set. Several scoring functions have been proposed over the past years, such as Shannon's entropy \cite{shannon1948mathematical}, variation ratio \cite{freeman1965elementary}, and Monte-Carlo (MC) dropout \cite{gal2016dropout}. Leveraging these scoring functions for the problem of data sample selection has been well studied for natural image classification \cite{gal2017deep} and semantic segmentation \cite{gorriz2017cost}. Nevertheless, it has been under explored for the problem of support set selection for few-shot microscopy image cell segmentation. In fact, to the best of our knowledge, we are the first to discuss a solution to this problem based on a new automatic selection mechanism to construct the support set of target microscopy image dataset that leverages consistency loss with respect to data augmentation as our scoring function. Our approach outperforms other selection functions, such as random, Shannon's entropy and MC-dropout approaches.

\section{Method}
In this section, we start by the problem definition followed by our support set selection method which consists of three steps. First, we propose our new self-supervised pretext task using the target dataset. Next, we utilise this self-supervised trained model together with our scoring function to select the images to include and annotate in the support set. Finally, we fine-tune the self-supervised trained model using the support set and then evaluate on the testing target images.
\subsection{Problem Definition}
Assume a collection of microscopy image datasets $\widehat{\mathcal{S}} = \{ \mathcal{S}_{i}\}_{i=1}^{|\widehat{\mathcal{S}}|}$. Each dataset is denoted by $\mathcal{S}_{i} =\{(\mathbf{x},\mathbf{y})_l\}_{l=1}^{|\mathcal{S}_i|}$, where $(\mathbf{x},\mathbf{y})_l$ is a pair of microscopy image $\mathbf{x} \in \mathcal{X} \subset \mathbb{R}^{H \times W}$ ($H \times W$ is the image size) and pixel-level binary cell segmentation ground-truth $\mathbf{y} \in \mathcal{Y} \subset \{0,1\}^{H \times W}$. All datasets in $\widehat{\mathcal{S}}$ are referred to as the sources. Note that each dataset represents a different microscopy image domain with different image appearance and cell segmentation task. We rely on the source datasets to learn a generic binary cell segmentation function $f_{\theta}:\mathcal{X} \to [0,1]^{H \times W}$ approximated by a deep neural network with parameters $\theta \in \Theta$. 

We also have the target dataset $\mathcal{T}= \{(\mathbf{x})_j\}_{j=1}^{|\mathcal{T}|}$, which contains images belonging to a different microscopy domain and a different cell segmentation task. Our main objective is to train a model to segment cells in microscopy images from the target set using a support set containing  annotated images denoted by $\mathcal{T}^{(|\mathcal{B}|)}$. The model is trained under the constraint of a limited annotation budget $|\mathcal{B}|$, i.e. few-shot training. Towards our objective, we consider a pool of unlabelled training images $\mathcal{T}^{(pool)} = \{(\mathbf{x})_n\}_{n=1}^{|\mathcal{T}^{(pool)}|}$, where $\mathcal{T}^{(|\mathcal{B}|)} \subset \mathcal{T}^{(pool)} \subset \mathcal{T}$, with the target testing set formed by $\widehat{\mathcal{T}} = \mathcal{T} \setminus \mathcal{T}^{(pool)}$. The samples in the support set are selected based on the pixel-wise binary cross entropy (BCE) scoring function $s:\mathcal{X} \times \Theta \to \mathbb{R}$ for assessing images in $\mathcal{T}^{(pool)}$. In summary, only the images $\mathbf{x} \in \mathcal{T}^{(pool)}$ that maximise the scoring function $s(.)$ are inserted into $\mathcal{T}^{(|\mathcal{B}|)}$ to be labelled by an oracle until the annotation budget is exhausted. Next we present our pretext task followed by our selection approach to construct $\mathcal{T}^{(|\mathcal{B}|)}$.

\subsection{Support Set Selection}

\label{PSL}

As previously mentioned, we assume that images belonging to the target testing set come from a different microscopy image domain and cell segmentation task, as the ones present in the source training set. Under this assumption, we argue that it is ineffective to use the source training set alone to train the segmentation model $f_{\theta}(.)$ to be used in the scoring function $s(.)$. Alternatively, we seek to adapt the entire model, i.e. the encoder and decoder, to the task of cell segmentation learning using a pseudo-labelled from the target data. In order to achieve this, we propose to employ classical computer vision operators to extract binary cell segmentation pseudo-labels for all the images in the target testing set. Specifically, we exploit image thresholding, global contrast enhancement using histogram equalisation \cite{pizer1987adaptive}, and dilation \cite{haralick1987image} filters to create a pseudo-labelled target dataset $\mathcal{T}^{(PL)} = \{(\mathbf{x},\mathbf{y}^{(PL)})_j \}_{j=1}^{|\mathcal{T}^{(PL)}|}$ where $\mathbf{y}^{(PL)}_j \in \mathcal{Y}$ is the corresponding pseudo-label of microscopy image $\mathbf{x}_j$. Note that we use all images of the target data $\mathcal{T}$ to create $\mathcal{T}^{(PL)}$. To acquire $\mathbf{y}^{(PL)}$, first, an input image $\mathbf{x}$ (from $\mathcal{T}^{(PL)}$) is passed through a histogram equalisation filter to get $\mathbf{x}^{(e)}$. Afterwards, we use a threshold filter $t_{thresh}(\mathbf{x}^{(e)}(\omega))$ defined as:
\begin{equation}
   \mathbf{x}^{(\gamma)}=t_{thresh}(\mathbf{x}^{(e)}(\omega)) = \{
        \begin{array}{ll}
            1, & \mathbf{x}^{(e)}(\omega) \leq \gamma \\
            0, & \mathbf{x}^{(e)}(\omega) > \gamma
        \end{array}
    , \forall \omega \in \Omega,
\end{equation}
where 
$\gamma \in \mathbb{R}$ is a threshold that approximately represents the pixel value of the cell of interest in the target data, and $\omega \in \Omega$ is a pixel address in the image lattice of size $H \times W$. 
At last, we apply a dilation filter of size $2\times2$ on the thresholded image $\mathbf{x}^{(\gamma)}$ to get the pseudo-label $\mathbf{y}^{(PL)}$. 

Our training  process starts with a meta-learning process that minimises the average BCE loss on the source training sets, where the loss for each training set $\mathcal{S}_i \in \widehat{\mathcal{S}}$ is defined as:
\begin{equation}\label{bce_pseudo}
    \begin{aligned}
     \mathcal{L}_{BCE} (\theta,\mathcal{S}_i) =
         - \frac{1}{|\mathcal{S}_i|}  \sum_{(\mathbf{x}, \mathbf{y})\in \mathcal{S}_i} \sum_{\omega \in \Omega }^{ }[\mathbf{w}(\omega) \mathbf{y}(\omega)\log( \mathbf{\hat{y}}(\omega) ) \\ +  
          (1-\mathbf{y}(\omega)) \log(1-\mathbf{\hat{y}} (\omega))],
\end{aligned}
\end{equation}
where  $\mathbf{\hat{y}}(\omega)$ denotes the prediction of the segmentation model (represented by $\mathbf{\hat{y}}=f_{\theta}(\mathbf{x})$) at spatial position $\omega$, and $\mathbf{w}(\omega)$ is a weighting factor defined by the ratio of foreground to background classes in the dataset.
After this initial training with~\eqref{bce_pseudo}, we fine-tune the parameters $\theta$ of our generic model $f_{\theta}(.)$ by minimising $\mathcal{L}_{BCE}(\theta, \mathcal{T}^{(PL)})$. By using $\mathcal{L}_{BCE}(\theta, \mathcal{T}^{(PL)})$, we seek to adapt the parameters of the generic model using the following optimisation:
\begin{equation}
    \label{optim:pseudo}
    \theta^{\prime} = \arg \min_{\theta} [\mathcal{L}_{BCE}(\theta,\mathcal{T}^{(PL)})].
\end{equation}

\begin{figure}[ht!]
    \centering
     \begin{tabular}{cc}
    \subfigure[$\mathbf{x}$]{\includegraphics[width=0.18\textwidth]{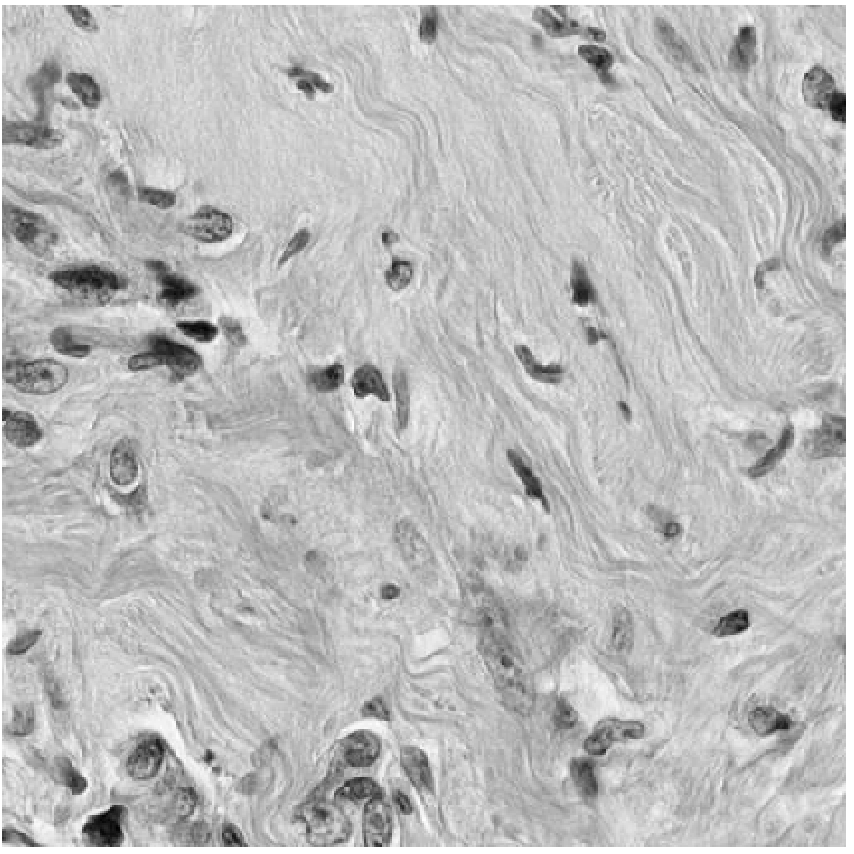}} 
    \subfigure[$\mathbf{x}^{(e)}$ ]{\includegraphics[width=0.18\textwidth]{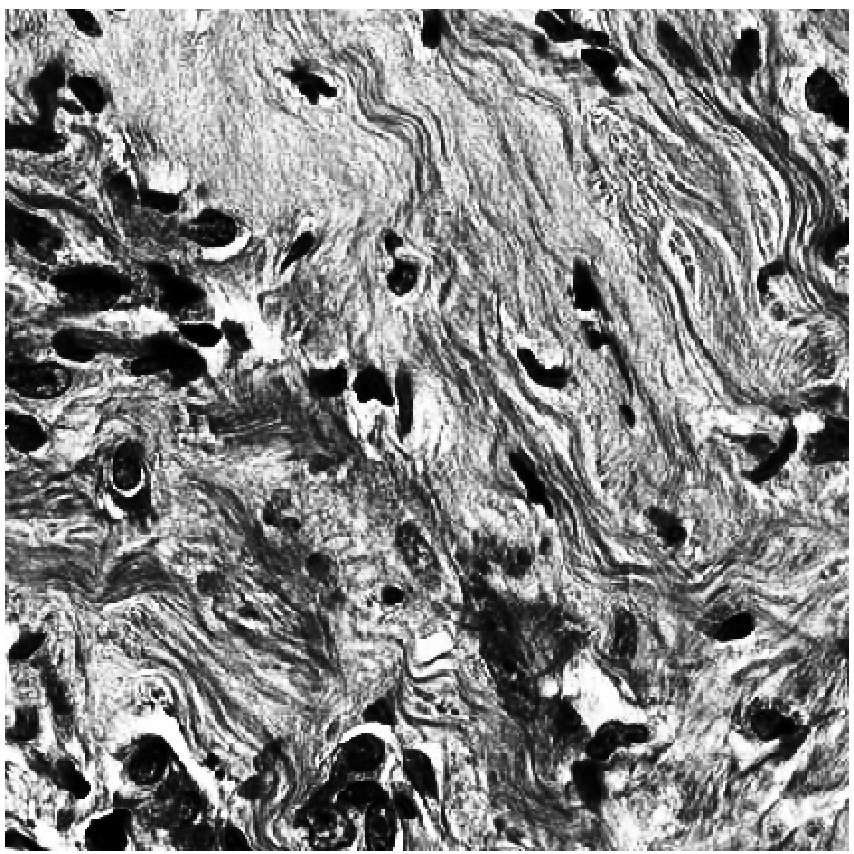}}
    \end{tabular}
    
     \begin{tabular}{cc}
   \subfigure[$\mathbf{x}^{(\gamma)}$]{\includegraphics[width=0.18\textwidth]{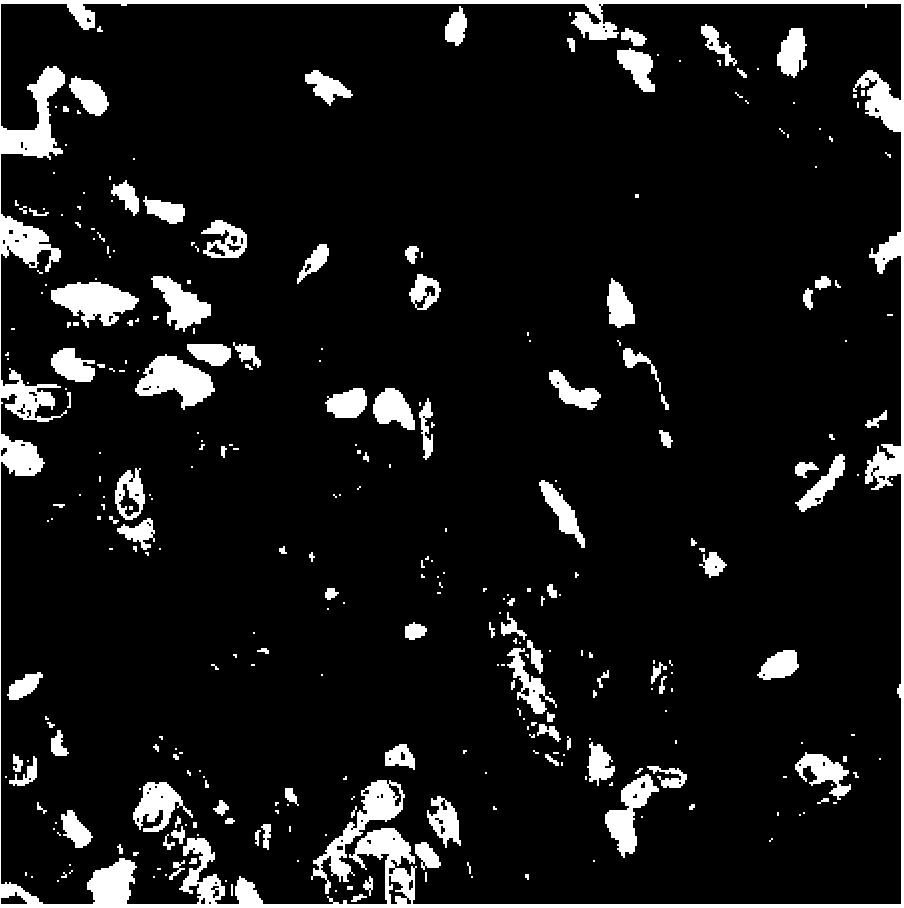}}
   \subfigure[$\mathbf{y}^{(PL)}$]{\includegraphics[width=0.18\textwidth]{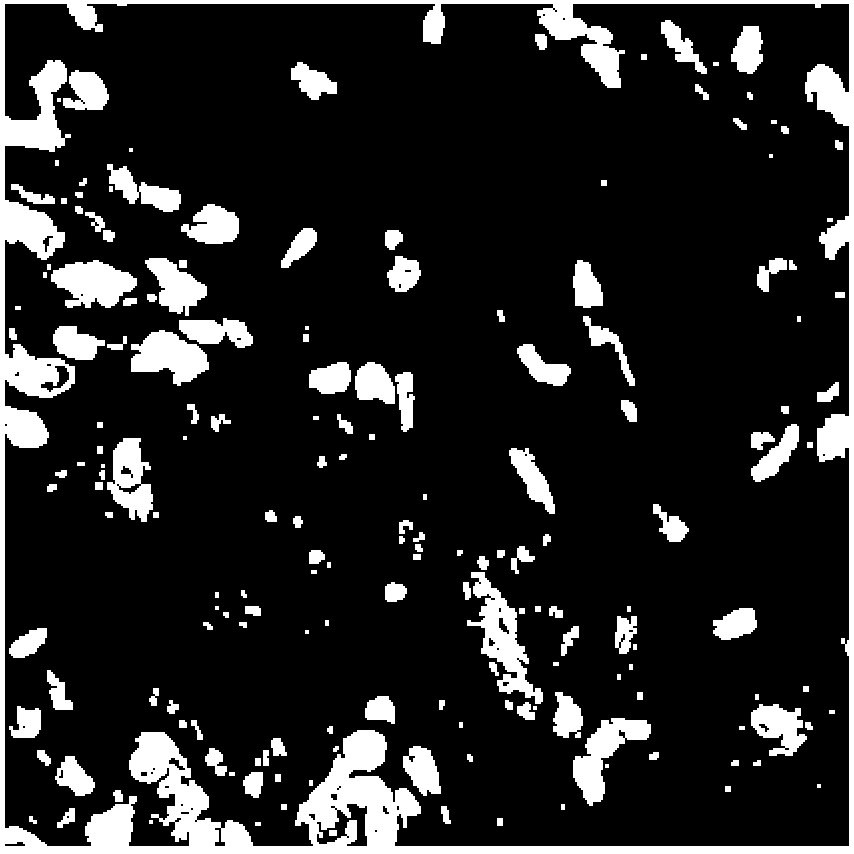}}
   \end{tabular}
  \caption{We present an example of our binary cell segmentation pseudo-label generation pipeline using a microscopy image $\mathbf{x}$ from TNBC target dataset \cite{naylor2018segmentation} in (a). First, we enhance the contrast of the image using histogram equalisation ($\mathbf{x}^{(e)}$) in (b), followed by  image thresholding ($\mathbf{x}^{(\gamma)}$) in (c) and dilation filtering in (d) to form the pseudo-label $\mathbf{y}^{(PL)}$.}
    \label{fig:Pseudolabelgen}
\end{figure}

Once the model is trained on pseudo-segmentation learning, we fix its learned parameters $\theta^{\prime}$ and start our selection process. In Algorithm \ref{algo}, we describe our complete support set selection approach. To begin with, each microscopy image $\mathbf{x}_{m} \in \mathcal{T}^{(pool)}, m \in \{1,2,\dots,|\mathcal{T}^{(pool)}|\}$ is augmented three times, each time with a different augmentation operation of magnitude $\psi$, namely, auto-contrast $C(\mathbf{x})$, brightness $B(\mathbf{x},\psi)$ and sharpness $S(\mathbf{x},\psi)$. We show in our experiments that these specific augmentations are suitable to our selection method for the microscopy image domain. Then, we calculate the BCE score for every microscopy image $\mathbf{x}$ in $\mathcal{T}^{(pool)}$ between the prediction of the augmented images, i.e. $\mathbf{\hat{y}}_{A} = f_{\theta^{\prime}}(A(\mathbf{x})), A(.) \in \{C(\mathbf{x}),B(\mathbf{x},\psi),S(\mathbf{x},\psi) \}$ and the prediction of the non-augmented image $\mathbf{\hat{y}} = f_{\theta}^{\prime}(\mathbf{x})$, as follows:
\begin{equation}\label{bce_selection}
\begin{aligned}
     s(\mathbf{x},\theta^{\prime})=
        -\sum_{A(.) \in \{C(\mathbf{x}),B(\mathbf{x},\psi),S(\mathbf{x},\psi) \}}  \frac{1}{{H \times W}}  \sum_{\omega \in \Omega }^{ } \\ [\mathbf{w}(\omega) \mathbf{\hat{y}}(\omega)\log( \mathbf{\hat{y}}_{A}(\omega) ) + 
         (1-\mathbf{\hat{y}}(\omega)) \log(1-\mathbf{\hat{y}}_{A} (\omega)) 
         ].
\end{aligned}
\end{equation}

The last step of the algorithm consists of forming the support set $\mathcal{T}^{(|\mathcal{B}|)}$ using the following optimisation:

\begin{equation}
\begin{split}
    \mathcal{T}^{(|\mathcal{B}|)} = & \arg \max_{\widehat{\mathcal{T}}^{(|\mathcal{B}|)} \subset \mathcal{T}^{(pool)}}  \sum_{\mathbf{x} \in \widehat{\mathcal{T}}^{(|\mathcal{B}|)}} s(\mathbf{x},\theta^{\prime}) \\
    & s.t. |\mathcal{T}^{(|\mathcal{B}|)}| = |\mathcal{B}|,
\end{split}
    \label{eq:selection_ss}
\end{equation}
where $|\mathcal{B}|$ is the size of the support set.
\subsection{Support Set Fine-tuning}
After selecting the support set images, we request an oracle (i.e., human expert) to manually annotate them. Next, we adapt $f_{\theta^{\prime}}(\mathbf{x})$ using $\mathcal{T}^{(|\mathcal{B}|)} \subset \mathcal{T}^{(pool)}$ and the binary cross entropy loss $\mathcal{L}_{BCE}(\theta^{\prime},\mathcal{T}^{(|\mathcal{B}|)})$, as in:
\begin{equation}
\label{finetune}
    \theta^{*} = \arg \min_{\theta^{\prime}} [\mathcal{L}_{BCE}(\theta^{\prime},\mathcal{T}^{(|\mathcal{B}|)})].
\end{equation}

We evaluate the fine-tuned model with $f_{\theta^{*}}(\mathbf{x})$ on the target test set $\widehat{\mathcal{T}} = \mathcal{T} \setminus \mathcal{T}^{(pool)}$.

\begin{algorithm}[!htbp]
\caption{Dynamic Support Set Selection based on Data Augmentation}\label{algo}
  \begin{algorithmic}[1]
  \STATE Input: Segmentation model $f_{\theta}(\mathbf{x})$ trained using $\mathcal{L}_{BCE}(\theta,\widehat{\mathcal{S}})$ from~\eqref{bce_pseudo}, pseudo-labelled target \\ dataset $\mathcal{T}^{(PL)} = \{(\mathbf{x},\mathbf{y}^{(PL)})_j \}_{j=1}^{|\mathcal{T}|}$,  unlabelled \\ target training set $\mathcal{T}^{(pool)}$, and annotation budget \\ $|\mathcal{B}|$
 
  \STATE Estimate $\theta^{\prime}$ from~\eqref{optim:pseudo} using $\mathcal{T}^{(PL)}$
  \FOR{$m$ = $1,2,\dots, |\mathcal{T}^{(pool)}|$}
     \STATE  Augment $\mathbf{x}_m \in \mathcal{T}^{(pool)}$ with auto-contrast \\ $C(\mathbf{x}_m)$, brightness $B(\mathbf{x}_m , \psi)$ and \\ sharpness $S(\mathbf{x}_m , \psi)$  
     \STATE Get prediction $\mathbf{\hat{y}}_m = f_{\theta^{\prime}}(\mathbf{x}_m)$
     \STATE Get predictions $\mathbf{\hat{y}}_{C,m} = f_{\theta^{\prime}}(C(\mathbf{x}_m))$, $\mathbf{\hat{y}}_{B,m} = f_{\theta^{\prime}}(B(\mathbf{x}_m , \psi))$, \\$\mathbf{\hat{y}}_{S,m} = f_{\theta^{\prime}}(S(\mathbf{x}_m , \psi))$
     \STATE Calculate score $s(\mathbf{x}_m,\theta^{\prime})$ from~\eqref{bce_selection} 
     
 \ENDFOR
 \STATE  Select $\mathcal{T}^{(|\mathcal{B}|)}$ from~\eqref{eq:selection_ss} using $\{s(\mathbf{x}_m,\theta^{\prime})\}_{m=1}^{|\mathcal{T}^{(pool)}|}$
 \STATE Estimate $\theta^*$ from~\eqref{finetune} using $\mathcal{T}^{(|\mathcal{B}|)}$

 \STATE Output: $f_{\theta^*}(.)$ to be tested on $\widehat{\mathcal{T}} = \mathcal{T} \setminus \mathcal{T}^{(pool)}$.
 \end{algorithmic}
 \end{algorithm}
\section{Experiments}
In our experiments, we use the same microscopy image cell segmentation benchmark of \cite{Dawoud2020}, which consists of five microscopy image datasets . More specifically, we have B5 and B39 datasets from the Broad Bioimage Benchmark Collection (BBBC)~\cite{lehmussola2007computational}. The former contains 1200 fluorescent synthetic stain cells images, while the latter contains 200 fluorescent synthetic stain cells. We also have Serial Section Transmission Electron Microscopy (ssTEM)~\cite{gerhard2013segmented} and Electron Microscopy (EM)~\cite{lucchi2013learning} datasets, containing 165 and 20 electron microscopy images, respectively, of mitochondria cells. The final dataset is Triple Negative Breast Cancer (TNBC) that consists of 50 histology images of breast biopsy~\cite{naylor2018segmentation}. We compare our scoring function against Shannon's entropy, MC-dropout, and random selection. Furthermore, we define our evaluation protocol since, to the best of our knowledge, we are the first to address support set selection problem.

\subsection{Implementation Details}
We employ the FCRN architecture \cite{xie2018microscopy} as our cell segmentation backbone model. First, we train the segmentation model $f_{\theta}(\mathbf{x})$ using the source datasets. However, instead of training the segmentation model from scratch, we exploit the readily trained model using a gradient-based meta-learning reptile algorithm provided by \cite{Dawoud2020}. 

\paragraph{Pseudo-label segmentation} We generate pseudo-labels using all target images as described in Sec. \ref{PSL}, where the threshold value $\gamma$ is heuristically defined per target dataset by visually inspecting the pixel values of cells of interest and setting the threshold value accordingly. Also, we augment $\mathcal{T}^{(PL)}$ by extracting image patches of size 256 $\times$ 256 from every microscopy image and its corresponding pseudo-label. We initialise the segmentation model with the parameter $\theta$ learned using source data, and train it for 100 epochs using $\mathcal{T}^{(PL)}$ and Adam optimiser \cite{kingma2014adam} with  learning rate 0.0001 and weight decay 0.0005. 

\paragraph{Support set selection} We experiment with support set sizes $|\mathcal{B}| \in \{1,3,5,7,10\}$ shots. For this stage, we noticed a better performance when selecting image patches instead of full resolution images. Every image in $\mathcal{T}^{(pool)}$ is then cropped to patches of size 256 $\times$ 256 pixels. The number of patches per image depends on the size of the original image, the crop step size, and the crop window size. We report the number of image patches per image in Table \ref{tab:patches}. Accordingly, we replace the support sizes mentioned earlier with the corresponding number of image patches per full resolution image times the number of shots. During the selection stage, we fix the trained model parameters $\theta^{\prime}$. For Shannon's entropy score, we do one forward pass for each image patch and calculate the prediction uncertainty using Shannon's entropy formula \cite{shannon1948mathematical}. As for the MC-dropout scoring approach, we observed that adding a dropout layer before the last convolutional layer yields the best results. Accordingly, we set the dropout probability to 0.5. We make ten forward passes for each image patch, average out the output probability distribution and then we calculate Shannon's entropy \cite{gal2016dropout}. Note that we only use MC-dropout during the selection stage, i.e., we do not train or fine-tune using MC-dropout following \cite{mi2019training}. For the random baseline, we randomly select image patches from $\mathcal{T}^{(pool)}$, where each image has an equal probability of selection. In our approach, we rely on auto-contrast, brightness, and sharpness augmentations provided in \cite{cubuk2020randaugment}. Moreover, we empirically set the magnitude of the distortion i.e. $\psi$ of brightness and sharpness augmentations to 1.3 \footnote{$\psi > 1$ means higher distortion.} for targets TNBC, EM, ssTEM and B39. As for B5, we increase the distortion magnitude to 1.6, otherwise, the added distortion would be insufficient and hence, the BCE loss values would be insignificant. The selection criterion is the same for Shannon's entropy, MC-dropout, and our approach, i.e., images that correspond to the top $|\mathcal{B}|$ loss scores are inserted into the support set.
\begin{table}[t!]
\begin{center}
\label{tab:patches}
\begin{tabular}{|c|c|c|c|c|c|}
\hline
Target        & TNBC & EM  & ssTEM & B5  & B39 \\ \hline
Patches/Image & 100  & 400 & 500   & 100 & 100 \\ \hline
\end{tabular}

\end{center}
\caption{Number of image patches extracted per full resolution image for each target dataset.}
\end{table}

\paragraph{Fine-tuning} We fine-tune the segmentation model $f_\theta{\prime}(\mathbf{x})$ using the resulting support set and Adam optimiser with learning rate 0.0001 and weight decay 0.0005 for 20 epochs. At last, we test the model using $\widehat{\mathcal{T}}$. Note that we use the test images in full resolution. 

\subsection{Evaluation Protocol}
Throughout our experiments, we rely on the mean intersection over union (mIoU) for quantifying the segmentation performance. We follow the same protocol as \cite{Dawoud2020} by conducting leave-one-dataset-out-cross-validation to split the microscopy datasets to source and target sets. In particular, we select four datasets as sources and treat the remaining dataset as target. The target set is randomly split to $\mathcal{T}^{(pool)}$ and $\widehat{\mathcal{T}}$. We repeat the random split ten times and report the mean and standard deviation of the mIoU over ten runs.
\subsection{Results and Discussion}

\begin{figure}[!ht]
    \centering
    \includegraphics[width=0.43\textwidth]{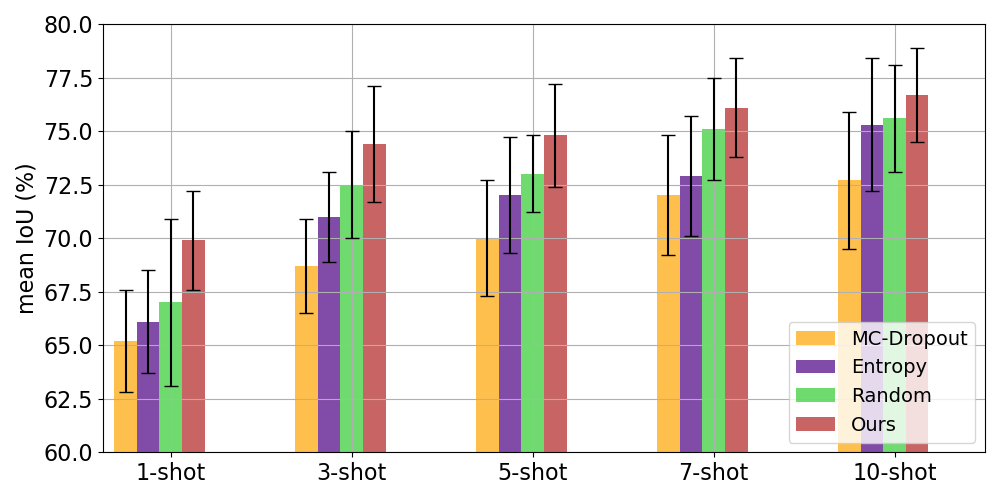}
    \caption{Comparison of mIoU results averaged over all datasets using our approach as well as Shannon's entropy, MC-dropout and random selection approaches. }
    \label{fig:avg_all_datasets}
\end{figure}

\begin{table*}[ht!]

\begin{center}
    
\begin{tabular}{|lccccc|}
\hline
\multicolumn{6}{|c|}{Target: TNBC}                                                                                                                                                  \\ \hline
\multicolumn{1}{|c|}{Method}     & \multicolumn{1}{c|}{\begin{tabular}[c]{@{}c@{}}1-shot\\ ($|\mathcal{B}|=100$)\end{tabular}} & \multicolumn{1}{c|}{\begin{tabular}[c]{@{}c@{}}3-shot\\ ($|\mathcal{B}|=300$)\end{tabular}} & \multicolumn{1}{c|}{\begin{tabular}[c]{@{}c@{}}5-shot\\ ($|\mathcal{B}|=500$)\end{tabular}} & \multicolumn{1}{c|}{\begin{tabular}[c]{@{}c@{}}7-shot\\ ($|\mathcal{B}|=700$)\end{tabular}} & \multicolumn{1}{c|}{\begin{tabular}[c]{@{}c@{}}10-shot\\ ($|\mathcal{B}|=1000$)\end{tabular}} \\ \hline
\multicolumn{1}{|c|}{Entropy}  & \multicolumn{1}{c|}{40.8\% ±3.9}                                                                            & \multicolumn{1}{c|}{44.1\% ±3.1}                                                                             & \multicolumn{1}{c|}{43.8\% ±4.7}                                                                             & \multicolumn{1}{c|}{45.6\% ±5.4}                                                               &           46.7\% ±5.8                                                                            \\ \hline
\multicolumn{1}{|c|}{MC-dropout} & \multicolumn{1}{c|}{40.8\% ±4.0}                                                                             & \multicolumn{1}{c|}{43.9\% ±3.1}                                                                            & \multicolumn{1}{c|}{43.7\% ±4.7}                                                                             & \multicolumn{1}{c|}{45.4\% ±5.4}                                                                             &     46.8\% ±5.8	                                                                                  \\ \hline
\multicolumn{1}{|c|}{Random} & \multicolumn{1}{c|}{37.0\% ±9.4}                                                               & \multicolumn{1}{c|}{44.7\% ±4.7}                                                               & \multicolumn{1}{c|}{	42.1\% ±2.3}                                                               & \multicolumn{1}{c|}{	45.7\% ±4.7}                                                               &      46.8\% ±4.8                                                                                 \\ \hline
\multicolumn{1}{|c|}{\textbf{\textit{Ours}}}  & \multicolumn{1}{c|}{\textbf{47.1\% ±3.5}}                                                       & \multicolumn{1}{c|}{\textbf{47.8\% ±5.8}}                                                     & \multicolumn{1}{c|}{\textbf{47.7\% ±5.0}}                                                     & \multicolumn{1}{c|}{\textbf{48.0\% ±5.2}}                                                                  &   \textbf{49.2\% ±4.9}                                                                          \\ \hline
\multicolumn{6}{|c|}{Target: EM}                                                                                                                                                  \\ \hline
\multicolumn{1}{|c|}{Method}     & \multicolumn{1}{c|}{\begin{tabular}[c]{@{}c@{}}1-shot\\ ($|\mathcal{B}|=400$)\end{tabular}} & \multicolumn{1}{c|}{\begin{tabular}[c]{@{}c@{}}3-shot\\ ($|\mathcal{B}|=1200$)\end{tabular}} & \multicolumn{1}{c|}{\begin{tabular}[c]{@{}c@{}}5-shot\\ ($|\mathcal{B}|=2000$)\end{tabular}} & \multicolumn{1}{c|}{\begin{tabular}[c]{@{}c@{}}7-shot\\ ($|\mathcal{B}|=2800$)\end{tabular}} & \multicolumn{1}{c|}{\begin{tabular}[c]{@{}c@{}}10-shot\\ ($|\mathcal{B}|=4000$)\end{tabular}} \\ \hline
\multicolumn{1}{|c|}{Entropy}  & \multicolumn{1}{c|}{61.0\% ±2.9}                                                                            & \multicolumn{1}{c|}{67.1\% ±2.7}                                                                             & \multicolumn{1}{c|}{68.6\% ±2.3}                                                                             & \multicolumn{1}{c|}{70.1\% ±1.9}                                                               &           72.9\% ±1.6                                                                            \\ \hline
\multicolumn{1}{|c|}{MC-dropout} & \multicolumn{1}{c|}{61.7\% ±2.6}                                                                             & \multicolumn{1}{c|}{63.4\% ±2.4}                                                                            & \multicolumn{1}{c|}{66.1\% ±2.1}                                                                             & \multicolumn{1}{c|}{68.0\% ±1.8}                                                                             &     70.0\% ±2.4	                                                                                  \\ \hline
\multicolumn{1}{|c|}{Random} & \multicolumn{1}{c|}{58.9\% ±5.0}                                                               & \multicolumn{1}{c|}{65.2\% ±2.9}                                                               & \multicolumn{1}{c|}{	68.6\% ±3.4}                                                               & \multicolumn{1}{c|}{	71.2\% ±2.9}                                                               &      72.2\% ±3.2                                                                                 \\ \hline
\multicolumn{1}{|c|}{\textbf{\textit{Ours}}}  & \multicolumn{1}{c|}{\textbf{62.0\% ±3.2}}                                                       & \multicolumn{1}{c|}{\textbf{69.8\% ±2.7}}                                                     & \multicolumn{1}{c|}{\textbf{70.6\% ±3.3}}                                                     & \multicolumn{1}{c|}{\textbf{73.1\% ±2.9}}                                                         &   \textbf{73.7\% ±3.2}                                                                        \\ \hline
\multicolumn{6}{|c|}{Target: ssTEM}                                                                                                                                                  \\ \hline

\multicolumn{1}{|c|}{Method}     & \multicolumn{1}{c|}{\begin{tabular}[c]{@{}c@{}}1-shot\\ ($|\mathcal{B}|=500$)\end{tabular}} & \multicolumn{1}{c|}{\begin{tabular}[c]{@{}c@{}}3-shot\\ ($|\mathcal{B}|=1500$)\end{tabular}} & \multicolumn{1}{c|}{\begin{tabular}[c]{@{}c@{}}5-shot\\ ($|\mathcal{B}|=2500$)\end{tabular}} & \multicolumn{1}{c|}{\begin{tabular}[c]{@{}c@{}}7-shot\\ ($|\mathcal{B}|=3200$)\end{tabular}} & \multicolumn{1}{c|}{\begin{tabular}[c]{@{}c@{}}10-shot\\ ($|\mathcal{B}|=5000$)\end{tabular}} \\ \hline
\multicolumn{1}{|c|}{Entropy}  & \multicolumn{1}{c|}{49.5\% ±3.6}                                                                            & \multicolumn{1}{c|}{62.6\% ±2.6}                                                                 & \multicolumn{1}{c|}{\textbf{65.4\% ±3.7}}                                                     & \multicolumn{1}{c|}{\textbf{67.7\% ±2.7}}                                                               &           68.6\% ±3.6                                                                            \\ \hline
\multicolumn{1}{|c|}{MC-dropout} & \multicolumn{1}{c|}{49.3\% ±3.0}                                                                             & \multicolumn{1}{c|}{62.2\% ±3.1}                                                              & \multicolumn{1}{c|}{64.6\% ±3.5}                                                                & \multicolumn{1}{c|}{67.9\% ±3.6}                                                                             &     \textbf{68.2\% ±4.5}	                                                                                  \\ \hline
\multicolumn{1}{|c|}{Random} & \multicolumn{1}{c|}{47.0\% ±4.8}                                                               & \multicolumn{1}{c|}{60.6\% ±4.0}                                                               & \multicolumn{1}{c|}{	63.1\% ±2.8}                                                               & \multicolumn{1}{c|}{	66.4\% ±4.1}                                                               &      67.1\% ±4.1                                                                                 \\ \hline
\multicolumn{1}{|c|}{\textbf{\textit{Ours}}}  & \multicolumn{1}{c|}{\textbf{51.3\% ±3.2}}                                                       & \multicolumn{1}{c|}{\textbf{63.3\% ±3.2}}                                                              & \multicolumn{1}{c|}{64.2\% ±3.2}                                                                   & \multicolumn{1}{c|}{67.3\% ±2.9}                                                                  &   68.7\% ±2.6                                                                          \\ \hline

\end{tabular}

\end{center}
\caption{mIoU results for the target testing sets of TNBC, EM, ssTEM. Best results are highlighted.}
\label{tab:TNBC}
\end{table*}

Figure \ref{fig:avg_all_datasets} shows a comparison of mIoU results averaged over all target datasets, where we observe a consistently better performance using our scoring function. Additionally, we report the numerical results of TNBC, EM and ssTEM in Tab. \ref{tab:TNBC}. Visual results are shown in Fig.~\ref{fig:VisualComp1}. For all support set sizes, we notice that our selection approach (\textbf{\textit{Ours}}) performs better than the random selection baseline at target cell segmentation tasks of complex microscopy domain involving different structural membranes other than the cell of interest, namely, histology (TNBC) and electron microscopy (ssTEM and EM). As for B5 and B39 (see supplementary material) we notice a high performance for both our approach and random selection, since both datasets come from a less complex microscopy domain consisting of synthetic cells and background only. As for MC-dropout and entropy approaches, we observe that our scoring function yields an overall better and more consistent performance in all target datasets. In very few cases, due to increasing support set size entropy and MC-dropout perform slightly better. We highlight the significance of our approach using Wilcoxon test (See Tab. 1 in supplementary material). Next, we conduct a series of ablation studies to analyse our approach under different configurations in Sec.~\ref{sec:ablation_study}.

\begin{figure}

    \centering
    \includegraphics[width=0.43\textwidth]{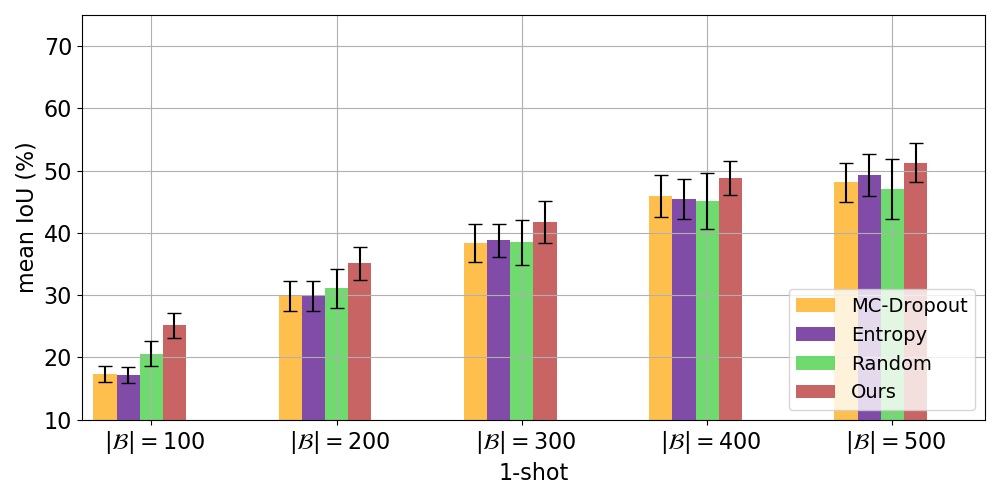}
    \caption{Impact on mIoU results when $|\mathcal{B}|$ extracting more image patches for 1-shot for target data ssTEM.}
\label{ab:ssTEM_SSS_1shot}
\end{figure}

\begin{figure*} [ht!]
    \centering
    \begin{tabular}{cc}
    \includegraphics[width=0.38\textwidth]{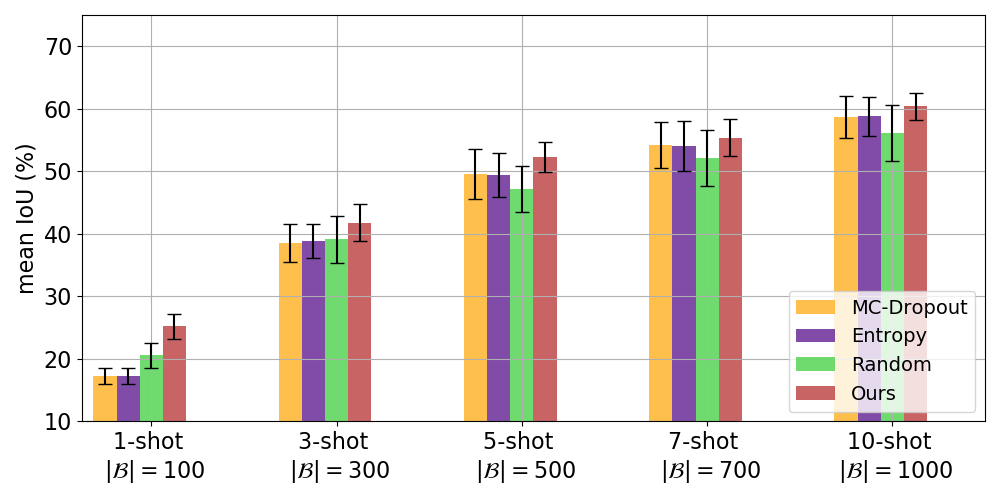}& 
    \includegraphics[width=0.38\linewidth]{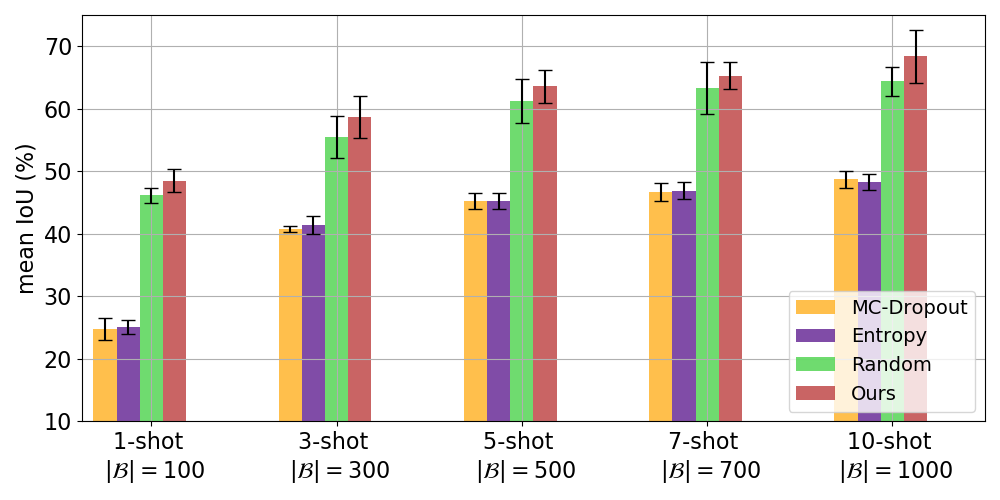} \\
      (a) ssTEM & (b) EM  \\
    \end{tabular}
     \caption{Impact of limiting $|\mathcal{B}|$ on the mIoU results for the target dataset (a) ssTEM and (b) EM. }
    \label{ab:ssTEM_EM_SSS}
\end{figure*}

\begin{figure*} [ht!]
    \centering
    \begin{tabular}{cc}
    \includegraphics[width=0.38\textwidth]{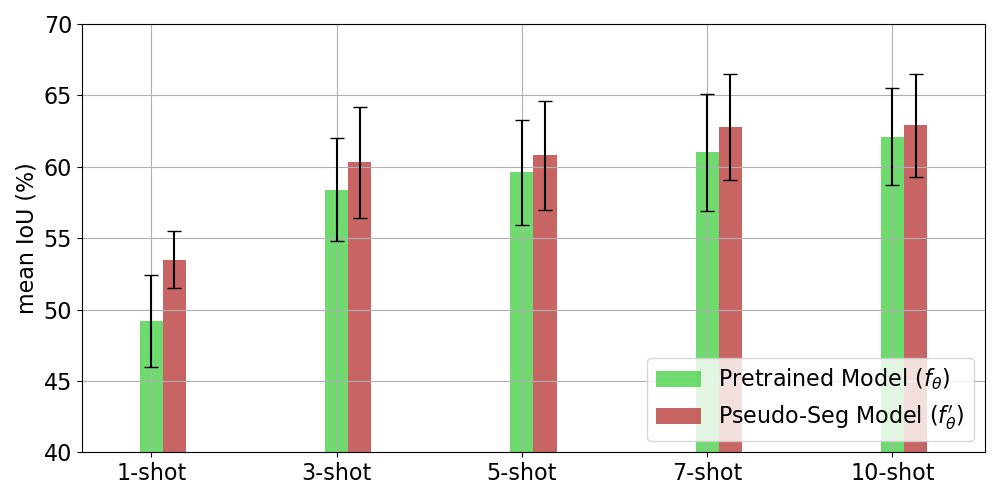}& 
    \includegraphics[width=0.38\linewidth]{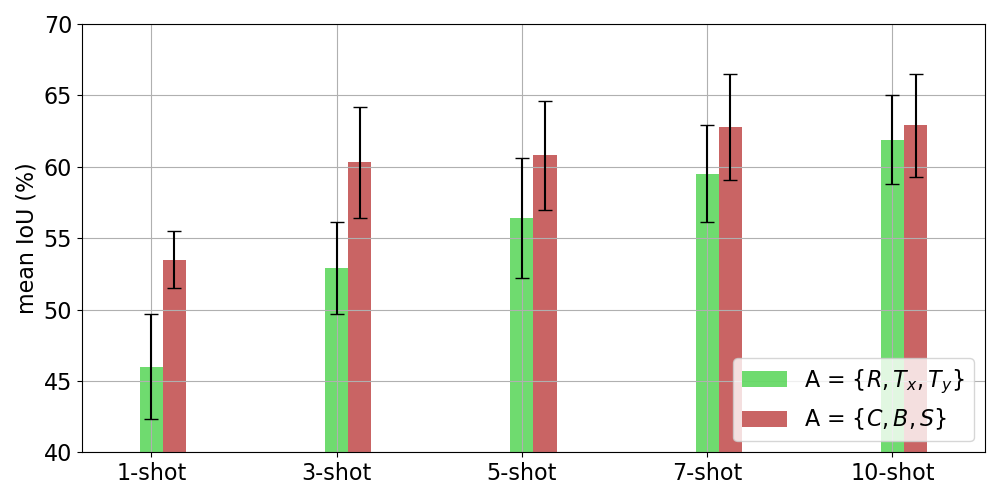} \\
      (a) Pseudo-label segmentation learning    & (b) Data augmentation  \\
    \end{tabular}
     \caption{Effect of the (a) pseudo-label segmentation learning (b) data augmentation on mIoU results averaged over target datasets.} 
    \label{fig:abPLFT_aug}
\end{figure*}
\begin{table*}[t!]

\label{tab:seg}

\begin{center}
\begin{tabular}{|c|c|c|c|c|c|}
\hline
Pseudo-label Generation                         & 1-shot & 3-shot & 5-shot & 7-shot & 10-shot \\ \hline

K-means &    54.4\% ±2.7  & 63.1\%  ±2.8   & 65.8\%  ±3.0 & 69.5\%  ±2.4 & 71.7\% ±2.6   \\ \hline
Equalisation+Threshold+Dilation &    56.7\% ±3.2& 66.6\%  ±3.0 &67.4\%  ±3.2   & 70.2\% ±2.9 &  71.2\% ±2.9   \\ \hline

\end{tabular}

\end{center}

\caption{Effect on mIoU results for using different pseudo-cell segmentation tasks averaged over two target datasets i.e. ssTEM and EM.}
\label{tab:pseudosegtask}
\end{table*}

\begin{table}[t!]
\centering
\begin{center}
\begin{tabular}{|c|c|c|c|c|c|}
\hline
Target                         & TNBC & EM & ssTEM & B5 & B39 \\ \hline
$f_{\theta^{\prime}}$ &    31±4.7  &  11±0.5  &   8±0.5&  99±0  &92±0.4     \\ \hline
\end{tabular}

\end{center}

\caption{Effect on mIoU results (\%) for testing the pseudo-label segmentation model $f_{\theta^{\prime}}(\mathbf{x})$ on the target datasets without support set fine-tuning.}
\label{tab:test}
\end{table}

\subsection{Ablation Study}
\label{sec:ablation_study}

\paragraph{Support set size.}
We examine the effect of limiting the support set size to only 100 image patches per target image. This only impacts the target datasets ssTEM and EM, while the results of the TNBC, B5 and B39 datasets remain unchanged. As reported in Fig. \ref{ab:ssTEM_EM_SSS}, we observe a significant performance drop in all selection approaches due to a smaller support set size in the fine-tuning process. On the other hand, our selection mechanism is still robust and yields better performance. This implies that our proposed scoring function remains more accurate compared to random, entropy and MC-dropout. We also report the results of using 200, 300, and 400 images per target image for dataset ssTEM in Fig. \ref{ab:ssTEM_SSS_1shot} which highlights the significant increase in performance as more image patches are extracted.

\paragraph{Pseudo-label cell segmentation learning and fine-tuning.} We claimed that the segmentation model $f_{\theta}(\mathbf{x})$ trained using source data only is not robust enough for scoring the target image predictions, which is the main motivation for the pretext task of pseudo-label cell segmentation learning. To support our claim, we conduct an experiment where we show the poorer performance of $f_{\theta}(\mathbf{x})$ in comparison to the pseudo-label segmentation model $f_{\theta^{\prime}}(\mathbf{x})$. The mIoU results in Fig.~\ref{fig:abPLFT_aug}  averaged over TNBC, EM and ssTEM target data sets show that our pretext task generally improves the performance due to better scoring and selection of support set. Numerical results for each target data set could be viewed in supplementary material. Moreover, we motivate the importance of support set fine-tuning by evaluating the pseudo-label segmentation model on the target test sets. The results in Table \ref{tab:test} clearly show that relying on pseudo-label segmentation learning alone is insufficient for targets TNBC, EM, and ssTEM, so it is necessary to perform support set fine-tuning with expert annotation of these target microscopy images. However, B5 and B39 yield fairly accurate results since their microscopy domain comprises only cells of interest and background, therefore, our pseudo-label segmentation pipeline results in a useful pseudo labels that are close to expert level annotation. Finally, we study a different technique to generate the pseudo-labels ($\mathbf{y}^{(PL)}$) for the task of pseudo-label cell segmentation learning. Namely, we use $K$-means \cite{pedregosa2011scikit} to cluster the pixels of each unlabelled target image to foreground and background classes i.e. binary segmentation map, hence, $K=2$. The study is conducted using target datasets EM and ssTEM. It can be noticed that the cells of interest (mitochondria) for both target datasets attain dark pixel values relative to other cells/membranes in the same image (see Fig. \ref{fig:VisualComp1}~a), therefore, pixels belonging to cluster center of darker pixel value are assigned to the foreground class while the pixels belonging to cluster center of brighter pixel value are treated as background. We report the results in Table~\ref{tab:pseudosegtask}. We clearly notice a better performance using the model fine-tuned on the pseudo-labels generated using the proposed pipeline in Sec. \ref{PSL} compared to the ones generated using $K$-means. However, the results also show that the support set selection algorithm is robust to the underlying pseudo-label generation technique and it can still perform well using both models fine-tuned to pseudo-labels of both techniques.

\paragraph{Data augmentation} Data augmentation may impact the consistency in the model predictions due to data distribution shifts. However, task specific augmentations may introduce a noticeable inconsistent predictions compared to random augmentations. In this work, we claim that pixel-level augmentations are more meaningful for the scoring function calculation than affine transformations. To this end, we compare our chosen augmentations with affine transformations, comprising image rotation (by 30°) and translation along the positive x (30$\%$ of H) and y axes (30$\%$ of W). We report our averaged mIoU results over TNBC, EM and ssTEM target data sets in Fig.~\ref{fig:abPLFT_aug}~b. Clearly, the proposed augmentations of contrast, brightness and sharpness have a positive impact in the BCE calculation, providing a better support set selection. Numerical results are listed in supplementary material.

\section{Conclusion}
We presented an approach to optimise support set selection for an effective fine-tuning process, hence, a better performance in few-shot microscopy image cell segmentation. Throughout our experiments, we demonstrated that by relying on our novel pretext task of pseudo-label cell segmentation learning and our scoring function, consistent and overall better results are achieved outperforming Shannon's entropy, MC-dropout and more importantly, random selection. Moreover, a series of ablation studies highlighted the important factors of our approach, which are the support set size, the impact of pseudo-label segmentation learning on support set selection, fine-tuning, and the effect of data augmentation on the selection process. Our work can be extended by combining other selection techniques such as diversity-based selection. Also, we plan to combine semi-supervised learning with support set fine-tuning.

\section*{Acknowledgments}
This work was partially funded by Deutsche Forschungsgemeinschaft (DFG), Research Training Group GRK 2203: Micro- and nano-scale sensor technologies for the lung (PULMOSENS),  by the German Federal Ministry for Economic Affairs and Energy within the project “KI Delta Learning” (Forderkennzeichen ¨
19A19013A) and the Australian Research Council through grant FT190100525.  G.C. acknowledges the support by the Alexander von Humboldt-Stiftung for the renewed research stay sponsorship.


{\small
\bibliographystyle{ieee_fullname}
\bibliography{references}

\begin{thebibliography}{10}\itemsep=-1pt

\bibitem{arteta2016counting}
Carlos Arteta, Victor Lempitsky, and Andrew Zisserman.
\newblock Counting in the wild.
\newblock In {\em European conference on computer vision}, pages 483--498.
  Springer, 2016.

\bibitem{baldi2012autoencoders}
Pierre Baldi.
\newblock Autoencoders, unsupervised learning, and deep architectures.
\newblock In {\em Proceedings of ICML workshop on unsupervised and transfer
  learning}, pages 37--49. JMLR Workshop and Conference Proceedings, 2012.

\bibitem{chen2019self}
Liang Chen, Paul Bentley, Kensaku Mori, Kazunari Misawa, Michitaka Fujiwara,
  and Daniel Rueckert.
\newblock Self-supervised learning for medical image analysis using image
  context restoration.
\newblock {\em Medical image analysis}, 58:101539, 2019.

\bibitem{chen2020simple}
Ting Chen, Simon Kornblith, Mohammad Norouzi, and Geoffrey Hinton.
\newblock A simple framework for contrastive learning of visual
  representations.
\newblock In {\em International conference on machine learning}, pages
  1597--1607. PMLR, 2020.

\bibitem{ciresan2012deep}
Dan Ciresan, Alessandro Giusti, Luca Gambardella, and J{\"u}rgen Schmidhuber.
\newblock Deep neural networks segment neuronal membranes in electron
  microscopy images.
\newblock {\em Advances in neural information processing systems},
  25:2843--2851, 2012.

\bibitem{cirecsan2013mitosis}
Dan~C Cire{\c{s}}an, Alessandro Giusti, Luca~M Gambardella, and J{\"u}rgen
  Schmidhuber.
\newblock Mitosis detection in breast cancer histology images with deep neural
  networks.
\newblock In {\em International conference on medical image computing and
  computer-assisted intervention}, pages 411--418. Springer, 2013.

\bibitem{cubuk2020randaugment}
Ekin~D Cubuk, Barret Zoph, Jonathon Shlens, and Quoc~V Le.
\newblock Randaugment: Practical automated data augmentation with a reduced
  search space.
\newblock In {\em Proceedings of the IEEE/CVF Conference on Computer Vision and
  Pattern Recognition Workshops}, pages 702--703, 2020.

\bibitem{dawoud2022edge}
Youssef Dawoud, Katharina Ernst, Gustavo Carneiro, and Vasileios Belagiannis.
\newblock Edge-based self-supervision for semi-supervised few-shot microscopy
  image cell segmentation.
\newblock In {\em International Workshop on Medical Optical Imaging and Virtual
  Microscopy Image Analysis}, pages 22--31. Springer, 2022.

\bibitem{Dawoud2020}
Youssef Dawoud, Julia Hornauer, Gustavo Carneiro, and Vasileios Belagiannis.
\newblock Few-shot microscopy image cell segmentation.
\newblock In {\em Machine Learning and Knowledge Discovery in Databases.
  Applied Data Science and Demo Track - European Conference, {ECML} {PKDD}
  2020, Ghent, Belgium, September 14-18, 2020, Proceedings, Part {V}}, volume
  12461, pages 139--154. Springer, 2020.

\bibitem{dijkstra2018centroidnet}
Klaas Dijkstra, J Loosdrecht, LRB Schomaker, and Marco~A Wiering.
\newblock Centroidnet: A deep neural network for joint object localization and
  counting.
\newblock In {\em Joint European Conference on Machine Learning and Knowledge
  Discovery in Databases}, pages 585--601. Springer, 2018.

\bibitem{faustino2009automatic}
Geisa~M Faustino, Marcelo Gattass, Stevens Rehen, and Carlos~JP de Lucena.
\newblock Automatic embryonic stem cells detection and counting method in
  fluorescence microscopy images.
\newblock In {\em 2009 IEEE International Symposium on Biomedical Imaging: From
  Nano to Macro}, pages 799--802. IEEE, 2009.

\bibitem{finn2017model}
Chelsea Finn, Pieter Abbeel, and Sergey Levine.
\newblock Model-agnostic meta-learning for fast adaptation of deep networks.
\newblock In {\em International Conference on Machine Learning}, pages
  1126--1135. PMLR, 2017.

\bibitem{freeman1965elementary}
Linton~C Freeman.
\newblock {\em Elementary applied statistics: for students in behavioral
  science}.
\newblock New York: Wiley, 1965.

\bibitem{gal2016dropout}
Yarin Gal and Zoubin Ghahramani.
\newblock Dropout as a bayesian approximation: Representing model uncertainty
  in deep learning.
\newblock In {\em international conference on machine learning}, pages
  1050--1059. PMLR, 2016.

\bibitem{gal2017deep}
Yarin Gal, Riashat Islam, and Zoubin Ghahramani.
\newblock Deep bayesian active learning with image data.
\newblock In {\em International Conference on Machine Learning}, pages
  1183--1192. PMLR, 2017.

\bibitem{gao2021utnet}
Yunhe Gao, Mu Zhou, and Dimitris~N Metaxas.
\newblock Utnet: a hybrid transformer architecture for medical image
  segmentation.
\newblock In {\em International Conference on Medical Image Computing and
  Computer-Assisted Intervention}, pages 61--71. Springer, 2021.

\bibitem{gerhard2013segmented}
Stephan Gerhard, Jan Funke, Julien Martel, Albert Cardona, and Richard Fetter.
\newblock Segmented anisotropic sstem dataset of neural tissue.
\newblock {\em figshare}, pages 0--0, 2013.

\bibitem{gidaris2018unsupervised}
Spyros Gidaris, Praveer Singh, and Nikos Komodakis.
\newblock Unsupervised representation learning by predicting image rotations.
\newblock {\em arXiv preprint arXiv:1803.07728}, 2018.

\bibitem{gorriz2017cost}
Marc Gorriz, Axel Carlier, Emmanuel Faure, and Xavier Giro-i Nieto.
\newblock Cost-effective active learning for melanoma segmentation.
\newblock {\em arXiv preprint arXiv:1711.09168}, 2017.

\bibitem{haralick1987image}
Robert~M Haralick, Stanley~R Sternberg, and Xinhua Zhuang.
\newblock Image analysis using mathematical morphology.
\newblock {\em IEEE transactions on pattern analysis and machine intelligence},
  (4):532--550, 1987.

\bibitem{jamaludin2017self}
Amir Jamaludin, Timor Kadir, and Andrew Zisserman.
\newblock Self-supervised learning for spinal mris.
\newblock In {\em Deep Learning in Medical Image Analysis and Multimodal
  Learning for Clinical Decision Support}, pages 294--302. Springer, 2017.

\bibitem{kingma2014adam}
Diederik~P Kingma and Jimmy Ba.
\newblock Adam: A method for stochastic optimization.
\newblock {\em arXiv preprint arXiv:1412.6980}, 2014.

\bibitem{lehmussola2007computational}
Antti Lehmussola, Pekka Ruusuvuori, Jyrki Selinummi, Heikki Huttunen, and Olli
  Yli-Harja.
\newblock Computational framework for simulating fluorescence microscope images
  with cell populations.
\newblock {\em IEEE transactions on medical imaging}, 26(7):1010--1016, 2007.

\bibitem{lu2015improved}
Zhi Lu, Gustavo Carneiro, and Andrew~P Bradley.
\newblock An improved joint optimization of multiple level set functions for
  the segmentation of overlapping cervical cells.
\newblock {\em IEEE Transactions on Image Processing}, 24(4):1261--1272, 2015.

\bibitem{lucchi2013learning}
Aur{\'e}lien Lucchi, Yunpeng Li, and Pascal Fua.
\newblock Learning for structured prediction using approximate subgradient
  descent with working sets.
\newblock In {\em Proceedings of the IEEE Conference on Computer Vision and
  Pattern Recognition}, pages 1987--1994, 2013.

\bibitem{makarevich2021metamedseg}
Anastasia Makarevich, Azade Farshad, Vasileios Belagiannis, and Nassir Navab.
\newblock Metamedseg: Volumetric meta-learning for few-shot organ segmentation.
\newblock {\em arXiv preprint arXiv:2109.09734}, 2021.

\bibitem{mi2019training}
Lu Mi, Hao Wang, Yonglong Tian, Hao He, and Nir Shavit.
\newblock Training-free uncertainty estimation for dense regression:
  Sensitivity as a surrogate.
\newblock {\em arXiv preprint arXiv:1910.04858}, 2019.

\bibitem{mondal2018few}
Arnab~Kumar Mondal, Jose Dolz, and Christian Desrosiers.
\newblock Few-shot 3d multi-modal medical image segmentation using generative
  adversarial learning.
\newblock {\em arXiv preprint arXiv:1810.12241}, 2018.

\bibitem{naylor2017nuclei}
Peter Naylor, Marick La{\'e}, Fabien Reyal, and Thomas Walter.
\newblock Nuclei segmentation in histopathology images using deep neural
  networks.
\newblock In {\em 2017 IEEE 14th international symposium on biomedical imaging
  (ISBI 2017)}, pages 933--936. IEEE, 2017.

\bibitem{naylor2018segmentation}
Peter Naylor, Marick La{\'e}, Fabien Reyal, and Thomas Walter.
\newblock Segmentation of nuclei in histopathology images by deep regression of
  the distance map.
\newblock {\em IEEE transactions on medical imaging}, 38(2):448--459, 2018.

\bibitem{nichol2018}
Alex Nichol, Joshua Achiam, and John Schulman.
\newblock On first-order meta-learning algorithms.
\newblock {\em CoRR}, abs/1803.02999, 2018.

\bibitem{noroozi2016unsupervised}
Mehdi Noroozi and Paolo Favaro.
\newblock Unsupervised learning of visual representations by solving jigsaw
  puzzles.
\newblock In {\em European conference on computer vision}, pages 69--84.
  Springer, 2016.

\bibitem{ouyang2020self}
Cheng Ouyang, Carlo Biffi, Chen Chen, Turkay Kart, Huaqi Qiu, and Daniel
  Rueckert.
\newblock Self-supervision with superpixels: Training few-shot medical image
  segmentation without annotation.
\newblock In {\em European Conference on Computer Vision}, pages 762--780.
  Springer, 2020.

\bibitem{pedregosa2011scikit}
Fabian Pedregosa, Ga{\"e}l Varoquaux, Alexandre Gramfort, Vincent Michel,
  Bertrand Thirion, Olivier Grisel, Mathieu Blondel, Peter Prettenhofer, Ron
  Weiss, Vincent Dubourg, et~al.
\newblock Scikit-learn: Machine learning in python.
\newblock {\em the Journal of machine Learning research}, 12:2825--2830, 2011.

\bibitem{pizer1987adaptive}
Stephen~M Pizer, E~Philip Amburn, John~D Austin, Robert Cromartie, Ari
  Geselowitz, Trey Greer, Bart ter Haar~Romeny, John~B Zimmerman, and Karel
  Zuiderveld.
\newblock Adaptive histogram equalization and its variations.
\newblock {\em Computer vision, graphics, and image processing},
  39(3):355--368, 1987.

\bibitem{ronneberger2015u}
Olaf Ronneberger, Philipp Fischer, and Thomas Brox.
\newblock U-net: Convolutional networks for biomedical image segmentation.
\newblock In {\em International Conference on Medical image computing and
  computer-assisted intervention}, pages 234--241. Springer, 2015.

\bibitem{settles2009active}
Burr Settles.
\newblock Active learning literature survey.
\newblock 2009.

\bibitem{shannon1948mathematical}
Claude~Elwood Shannon.
\newblock A mathematical theory of communication.
\newblock {\em The Bell system technical journal}, 27(3):379--423, 1948.

\bibitem{taleb20203d}
Aiham Taleb, Winfried Loetzsch, Noel Danz, Julius Severin, Thomas Gaertner,
  Benjamin Bergner, and Christoph Lippert.
\newblock 3d self-supervised methods for medical imaging, 2020.

\bibitem{wahlby2004combining}
Carolina W{\"a}hlby, I-M Sintorn, Fredrik Erlandsson, Gunilla Borgefors, and
  Ewert Bengtsson.
\newblock Combining intensity, edge and shape information for 2d and 3d
  segmentation of cell nuclei in tissue sections.
\newblock {\em Journal of microscopy}, 215(1):67--76, 2004.

\bibitem{xie2018microscopy}
Weidi Xie, J~Alison Noble, and Andrew Zisserman.
\newblock Microscopy cell counting and detection with fully convolutional
  regression networks.
\newblock {\em Computer methods in biomechanics and biomedical engineering:
  Imaging \& Visualization}, 6(3):283--292, 2018.

\bibitem{xing2016robust}
Fuyong Xing and Lin Yang.
\newblock Robust nucleus/cell detection and segmentation in digital pathology
  and microscopy images: a comprehensive review.
\newblock {\em IEEE reviews in biomedical engineering}, 9:234--263, 2016.

\bibitem{zhang2012classifying}
Xianglilan Zhang, Hongnan Wang, Tony~J Collins, Zhigang Luo, and Ming Li.
\newblock Classifying stem cell differentiation images by information distance.
\newblock In {\em Joint European Conference on Machine Learning and Knowledge
  Discovery in Databases}, pages 269--282. Springer, 2012.

\end{thebibliography}
}

\onecolumn

\appendix
\renewcommand*\appendixpagename{\Large Supplementary Material}
\appendixpage


In the supplementary material we present the Wilcoxon significance test of our approach against random, MC-dropout, and entropy baselines in Tab. \ref{tab:sig}. Our null hypothesis ($H_0$) assumes that the mIoU of the baseline is greater than our approach i.e. $\mathrm{mIoU}_{baseline} > \mathrm{mIoU}_{Ours}$. On the other hand, the alternate ($H_a$) claims the opposite i.e. $\mathrm{mIoU}_{baseline} < \mathrm{mIoU}_{Ours}$. We use significance level of 0.05. The results clearly demonstrate the significance of our approach in many cases (36 out of 75) relative to the baselines which supports the alternate hypothesis. We show additional visual results for B5 and B39 datasets and their corresponding numerical results in Tab. \ref{tab:B39_B5}. Furthermore, we report the numerical results of our ablation studies in Tab. \ref{ab:ssTEM_SSS} to \ref{tab:aug}. First, in Tab. \ref{ab:ssTEM_SSS} and \ref{ab:EM_SSS} we show the impact of limiting the image patches extracted per target image for ssTEM and EM target datasets respectively. Second, we present the impact of increasing the number of patches extracted per 1-shot for ssTEM target dataset in Tab. \ref{ab:ssTEM_SSS}. Afterwards, in Tab. \ref{tab:PLFT} we report the performance on target datasets TNBC, EM, and ssTEM between using the pretrained model and the model trained using pseudo-label segmentation learning for support set selection and fine-tuning. In Tab. \ref{tab:aug} we report the results of using pixel-level augmentations compared to affine augmentations. Finally, we report the fine-tuning performance of human-expert selection compared to our selection approach in Tab. \ref{tab:expert}.

\begin{table*}[ht!]
\begin{center}
    
\begin{tabular}{|lccccc|}
\hline
\multicolumn{6}{|c|}{Target: TNBC}                                                                                                                                                  \\ \hline
\multicolumn{1}{|c|}{Method}     & \multicolumn{1}{c|}{\begin{tabular}[c]{@{}c@{}}1-shot\\ ($|\mathcal{B}|=100$)\end{tabular}} & \multicolumn{1}{c|}{\begin{tabular}[c]{@{}c@{}}3-shot\\ ($|\mathcal{B}|=300$)\end{tabular}} & \multicolumn{1}{c|}{\begin{tabular}[c]{@{}c@{}}5-shot\\ ($|\mathcal{B}|=500$)\end{tabular}} & \multicolumn{1}{c|}{\begin{tabular}[c]{@{}c@{}}7-shot\\ ($|\mathcal{B}|=700$)\end{tabular}} & \multicolumn{1}{c|}{\begin{tabular}[c]{@{}c@{}}10-shot\\ ($|\mathcal{B}|=1000$)\end{tabular}} \\ \hline
\multicolumn{1}{|c|}{Entropy}  & \multicolumn{1}{c|}{0.003}                                                                            & \multicolumn{1}{c|}{0.023}                                                                             & \multicolumn{1}{c|}{0.062}                                                                             & \multicolumn{1}{c|}{0.037}                                                               &   0.023                                                                              \\ \hline
\multicolumn{1}{|c|}{MC-dropout} & \multicolumn{1}{c|}{0.003}                                                                             & \multicolumn{1}{c|}{0.023}                                                                            & \multicolumn{1}{c|}{0.063}                                                                             & \multicolumn{1}{c|}{0.018}                                                                             &     0.023                                                                                  \\ \hline
\multicolumn{1}{|c|}{Random} & \multicolumn{1}{c|}{0.008}                                                               & \multicolumn{1}{c|}{0.011}                                                               & \multicolumn{1}{c|}{0.011}                                                               & \multicolumn{1}{c|}{0.037}                                                               &    0.003                                                                                   \\ \hline

\multicolumn{6}{|c|}{Target: EM}                                                                                                                                                  \\ \hline
\multicolumn{1}{|c|}{Method}     & \multicolumn{1}{c|}{\begin{tabular}[c]{@{}c@{}}1-shot\\ ($|\mathcal{B}|=400$)\end{tabular}} & \multicolumn{1}{c|}{\begin{tabular}[c]{@{}c@{}}3-shot\\ ($|\mathcal{B}|=1200$)\end{tabular}} & \multicolumn{1}{c|}{\begin{tabular}[c]{@{}c@{}}5-shot\\ ($|\mathcal{B}|=2000$)\end{tabular}} & \multicolumn{1}{c|}{\begin{tabular}[c]{@{}c@{}}7-shot\\ ($|\mathcal{B}|=2800$)\end{tabular}} & \multicolumn{1}{c|}{\begin{tabular}[c]{@{}c@{}}10-shot\\ ($|\mathcal{B}|=4000$)\end{tabular}} \\ \hline
\multicolumn{1}{|c|}{Entropy}  & \multicolumn{1}{c|}{0.25}                                                                            & \multicolumn{1}{c|}{0.003}                                                                             & \multicolumn{1}{c|}{0.046}                                                                             & \multicolumn{1}{c|}{0.006}                                                               &        0.14                                                                              \\ \hline
\multicolumn{1}{|c|}{MC-dropout} & \multicolumn{1}{c|}{0.29}                                                                             & \multicolumn{1}{c|}{0.003}                                                                            & \multicolumn{1}{c|}{0.008}                                                                             & \multicolumn{1}{c|}{0.003}                                                                             &     0.003	                                                                                  \\ \hline
\multicolumn{1}{|c|}{Random} & \multicolumn{1}{c|}{0.08}                                                               & \multicolumn{1}{c|}{0.005}                                                               & \multicolumn{1}{c|}{0.057}                                                               & \multicolumn{1}{c|}{	0.057}                                                               &      0.029                                                                               \\ \hline

\multicolumn{6}{|c|}{Target: ssTEM}                                                                                                                                                  \\ \hline

\multicolumn{1}{|c|}{Method}     & \multicolumn{1}{c|}{\begin{tabular}[c]{@{}c@{}}1-shot\\ ($|\mathcal{B}|=500$)\end{tabular}} & \multicolumn{1}{c|}{\begin{tabular}[c]{@{}c@{}}3-shot\\ ($|\mathcal{B}|=1500$)\end{tabular}} & \multicolumn{1}{c|}{\begin{tabular}[c]{@{}c@{}}5-shot\\ ($|\mathcal{B}|=2500$)\end{tabular}} & \multicolumn{1}{c|}{\begin{tabular}[c]{@{}c@{}}7-shot\\ ($|\mathcal{B}|=3200$)\end{tabular}} & \multicolumn{1}{c|}{\begin{tabular}[c]{@{}c@{}}10-shot\\ ($|\mathcal{B}|=5000$)\end{tabular}} \\ \hline
\multicolumn{1}{|c|}{Entropy}  & \multicolumn{1}{c|}{0.037}                                                                            & \multicolumn{1}{c|}{0.08}                                                                 & \multicolumn{1}{c|}{0.96}                                                     & \multicolumn{1}{c|}{0.89}                                                               &  0.36                                                                                 \\ \hline
\multicolumn{1}{|c|}{MC-dropout} & \multicolumn{1}{c|}{0.03}                                                                             & \multicolumn{1}{c|}{0.12}                                                              & \multicolumn{1}{c|}{0.81}                                                                & \multicolumn{1}{c|}{0.94}                                                                             &     0.71	                                                                                  \\ \hline
\multicolumn{1}{|c|}{Random} & \multicolumn{1}{c|}{0.011}                                                               & \multicolumn{1}{c|}{0.006}                                                               & \multicolumn{1}{c|}{0.14}                                                               & \multicolumn{1}{c|}{0.52}                                                               &      0.084                                                                                \\ \hline

\multicolumn{6}{|c|}{Target: B39}                                                                                                                                                  \\ \hline

\multicolumn{1}{|c|}{Method}     & \multicolumn{1}{c|}{\begin{tabular}[c]{@{}c@{}}1-shot\\ ($|\mathcal{B}|=100$)\end{tabular}} & \multicolumn{1}{c|}{\begin{tabular}[c]{@{}c@{}}3-shot\\ ($|\mathcal{B}|=300$)\end{tabular}} & \multicolumn{1}{c|}{\begin{tabular}[c]{@{}c@{}}5-shot\\ ($|\mathcal{B}|=500$)\end{tabular}} & \multicolumn{1}{c|}{\begin{tabular}[c]{@{}c@{}}7-shot\\ ($|\mathcal{B}|=700$)\end{tabular}} & \multicolumn{1}{c|}{\begin{tabular}[c]{@{}c@{}}10-shot\\ ($|\mathcal{B}|=1000$)\end{tabular}} \\ \hline
\multicolumn{1}{|c|}{Entropy}  & \multicolumn{1}{c|}{0.003}                                                                    & \multicolumn{1}{c|}{0.003}                                                                & \multicolumn{1}{c|}{0.003}                                                                             & \multicolumn{1}{c|}{0.003}                                                               &0.003                                                                                \\ \hline
\multicolumn{1}{|c|}{MC-dropout} & \multicolumn{1}{c|}{0.003}                                                                             & \multicolumn{1}{c|}{0.003}                                                                            & \multicolumn{1}{c|}{0.003}                                                                             & \multicolumn{1}{c|}{0.003}                                                                             &    0.003                                                                               \\ \hline
\multicolumn{1}{|c|}{Random} & \multicolumn{1}{c|}{0.99}                                                             & \multicolumn{1}{c|}{0.98}                                                    & \multicolumn{1}{c|}{0.99}                                                               & \multicolumn{1}{c|}{0.99}                                                               &  0.99                                                                              \\ \hline

\multicolumn{6}{|c|}{Target: B5}                                                                                                                                                  \\ \hline
\multicolumn{1}{|c|}{Method}     & \multicolumn{1}{c|}{\begin{tabular}[c]{@{}c@{}}1-shot\\ ($|\mathcal{B}|=100$)\end{tabular}} & \multicolumn{1}{c|}{\begin{tabular}[c]{@{}c@{}}3-shot\\ ($|\mathcal{B}|=300$)\end{tabular}} & \multicolumn{1}{c|}{\begin{tabular}[c]{@{}c@{}}5-shot\\ ($|\mathcal{B}|=500$)\end{tabular}} & \multicolumn{1}{c|}{\begin{tabular}[c]{@{}c@{}}7-shot\\ ($|\mathcal{B}|=700$)\end{tabular}} & \multicolumn{1}{c|}{\begin{tabular}[c]{@{}c@{}}10-shot\\ ($|\mathcal{B}|=1000$)\end{tabular}} \\ \hline
\multicolumn{1}{|c|}{Entropy}  & \multicolumn{1}{c|}{0.99}                                                                    & \multicolumn{1}{c|}{0.99}                                                                & \multicolumn{1}{c|}{0.99}                                                                             & \multicolumn{1}{c|}{0.99}                                                               &           0.99                                                                            \\ \hline
\multicolumn{1}{|c|}{MC-dropout} & \multicolumn{1}{c|}{0.99}                                                                             & \multicolumn{1}{c|}{0.99}                                                                            & \multicolumn{1}{c|}{0.99}                                                                             & \multicolumn{1}{c|}{0.99}                                                                             &     0.99	                                                                                  \\ \hline
\multicolumn{1}{|c|}{Random} & \multicolumn{1}{c|}{0.99}                                                             & \multicolumn{1}{c|}{0.99}                                                        & \multicolumn{1}{c|}{0.99}                                                              & \multicolumn{1}{c|}{0.99}                                                            &      0.99                                                                                \\ \hline
\end{tabular}

\end{center}
\caption{Wilcoxon signed-rank significance test results of Ours vs $\{$ Entropy, MC-dropout, Random $\}$. Our null hypothesis $H_0$ assumes that the mIoU of the baseline is greater than our approach. While the alternate hypothesis $H_a$ claims the opposite. We set the significance level at 0.05 i.e. we reject $H_0$ if $p < 0.05$ . We report the $p$-values for targets TNBC, EM, ssTEM, B39, and B5.}
\label{tab:sig}
\end{table*}

\begin{figure*}[h!]
\centering
	
    \rotatebox[origin=l]{90}{B5}
	\includegraphics[width=3.9cm]{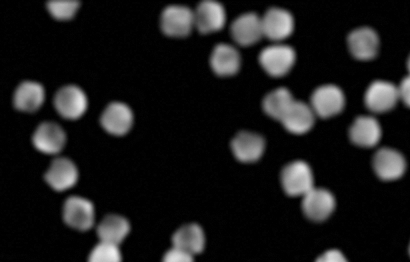}
	\includegraphics[width=3.9cm]{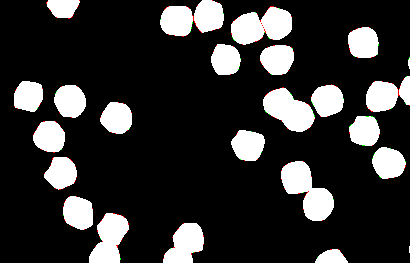}
	\includegraphics[width=3.9cm]{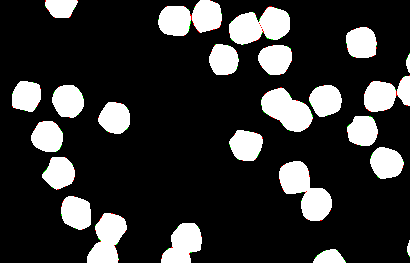}

	\rotatebox[origin=l]{90}{B39}
	\subfigure[c][Input]{\includegraphics[width=3.9cm]{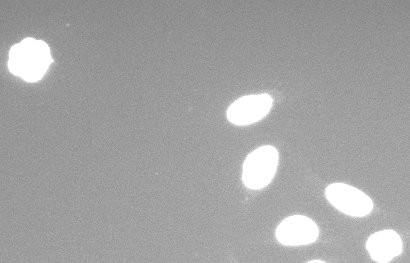}}
	\subfigure[c][Random]{\includegraphics[width=3.9cm]{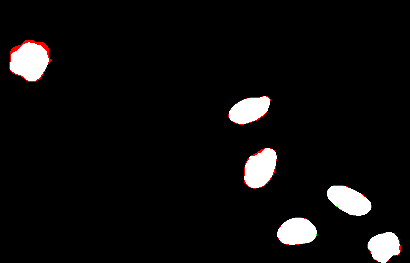}}
	\subfigure[c][Ours]{\includegraphics[width=3.9cm]{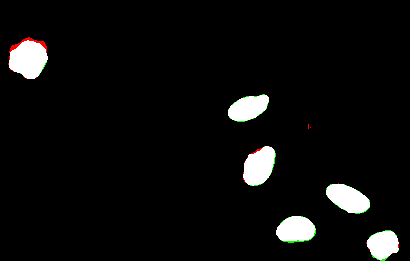}}

	\caption{Visual Result. We visually compare our scoring function (Ours) to random selection using the FCRN architecture at $|\mathcal{B}| = 3$-shots.  The red colour corresponds to false positive, the green colour to false negative, the black colour to true negative, and the white colour to true positive. Best viewed in colour.}
	\label{fig:VisualComp1}
\end{figure*}

\begin{table*}[t!]

\begin{center}
\begin{tabular}{|lccccc|}
\hline
\multicolumn{6}{|c|}{Target: TNBC}                                                                                                                                                  \\ \hline
\multicolumn{1}{|c|}{Model}     & \multicolumn{1}{c|}{\begin{tabular}[c]{@{}c@{}}1-shot\\ \end{tabular}} & \multicolumn{1}{c|}{\begin{tabular}[c]{@{}c@{}}3-shot\\ \end{tabular}} & \multicolumn{1}{c|}{\begin{tabular}[c]{@{}c@{}}5-shot\\ \end{tabular}} & \multicolumn{1}{c|}{\begin{tabular}[c]{@{}c@{}}7-shot\\ \end{tabular}} & \multicolumn{1}{c|}{\begin{tabular}[c]{@{}c@{}}10-shot\\ \end{tabular}} \\ \hline

\multicolumn{1}{|c|}{$f_{\theta}(\mathbf{x})$}  & \multicolumn{1}{c|}{44.5\% ±5.0}                                                                  & \multicolumn{1}{c|}{	47.8\% ±6.2}                                                               & \multicolumn{1}{c|}{47.7\% ±7.5}                                                                   & \multicolumn{1}{c|}{48.3\% ±7.0}                                                                  &   48.7\% ±6.2                                                                          \\ \hline

\multicolumn{1}{|c|}{$f_{\theta^\prime}(\mathbf{x})$}  & \multicolumn{1}{c|}{47.1\% ±3.5}                                                                  & \multicolumn{1}{c|}{	47.8\% ±5.8}                                                               & \multicolumn{1}{c|}{47.7\% ±5.0}                                                             & \multicolumn{1}{c|}{48.0\% ±5.2}                                                                  &   49.2\% ±4.9                                                                          \\ \hline
\multicolumn{6}{|c|}{Target: EM}  \\ \hline
\multicolumn{1}{|c|}{Model}     & \multicolumn{1}{c|}{\begin{tabular}[c]{@{}c@{}}1-shot\\ \end{tabular}} & \multicolumn{1}{c|}{\begin{tabular}[c]{@{}c@{}}3-shot\\ \end{tabular}} & \multicolumn{1}{c|}{\begin{tabular}[c]{@{}c@{}}5-shot\\ \end{tabular}} & \multicolumn{1}{c|}{\begin{tabular}[c]{@{}c@{}}7-shot\\ \end{tabular}} & \multicolumn{1}{c|}{\begin{tabular}[c]{@{}c@{}}10-shot\\ \end{tabular}} \\ \hline
\multicolumn{1}{|c|}{$f_{\theta}(\mathbf{x})$}  & \multicolumn{1}{c|}{52.7\% ±1.9}                                                                  & \multicolumn{1}{c|}{	64.8\% ±1.7}                                                               & \multicolumn{1}{c|}{66.4\% ±2.0}                                                                   & \multicolumn{1}{c|}{68.0\% ±2.0}                                                                  &   70.4\% ±1.0                                                                          \\ \hline

\multicolumn{1}{|c|}{$f_{\theta^\prime}(\mathbf{x})$}  &  \multicolumn{1}{c|}{62.0\% ±3.2}                                                                  & \multicolumn{1}{c|}{	69.8\% ±2.7}                                                               & \multicolumn{1}{c|}{70.6\% ±3.3}                                                                   & \multicolumn{1}{c|}{73.1\% ±2.9}                                                                  &   73.7\% ±3.2                                                                            \\ \hline

\multicolumn{6}{|c|}{Target: ssTEM}  \\ \hline
\multicolumn{1}{|c|}{Model}     & \multicolumn{1}{c|}{\begin{tabular}[c]{@{}c@{}}1-shot\\ \end{tabular}} & \multicolumn{1}{c|}{\begin{tabular}[c]{@{}c@{}}3-shot\\ \end{tabular}} & \multicolumn{1}{c|}{\begin{tabular}[c]{@{}c@{}}5-shot\\ \end{tabular}} & \multicolumn{1}{c|}{\begin{tabular}[c]{@{}c@{}}7-shot\\ \end{tabular}} & \multicolumn{1}{c|}{\begin{tabular}[c]{@{}c@{}}10-shot\\ \end{tabular}} \\ \hline

\multicolumn{1}{|c|}{$f_{\theta}(\mathbf{x})$}  & \multicolumn{1}{c|}{50.6\% ±2.8}                                                                  & \multicolumn{1}{c|}{	62.8\% ±3.0}                                                               & \multicolumn{1}{c|}{64.8\% ±1.7}                                                                   & \multicolumn{1}{c|}{66.7\% ±3.4}                                                                  &   67.2\% ±3.0                                                                          \\ \hline

\multicolumn{1}{|c|}{$f_{\theta^\prime}(\mathbf{x})$}  &  \multicolumn{1}{c|}{51.3\% ±3.2}                                                                  & \multicolumn{1}{c|}{63.3\% ±3.2}                                                               & \multicolumn{1}{c|}{64.2\% ±3.2}                                                                   & \multicolumn{1}{c|}{67.3\% ±2.9}                                                                  &   68.7\% ±2.6                                                                            \\ \hline

\end{tabular}

\end{center}

\caption{Effect on mIoU results of the pseudo-label segmentation learning and selection on the target data sets TNBC, EM, and ssTEM.}
\label{tab:PLFT}
\end{table*}

\

\begin{table*}[t!]


\begin{center}

\begin{tabular}{|lccccc|}
\hline
\multicolumn{6}{|c|}{Target: TNBC}                                                                                                                                                  \\ \hline
\multicolumn{1}{|c|}{Augmentation}     & \multicolumn{1}{c|}{\begin{tabular}[c]{@{}c@{}}1-shot\\ \end{tabular}} & \multicolumn{1}{c|}{\begin{tabular}[c]{@{}c@{}}3-shot\\ \end{tabular}} & \multicolumn{1}{c|}{\begin{tabular}[c]{@{}c@{}}5-shot\\ \end{tabular}} & \multicolumn{1}{c|}{\begin{tabular}[c]{@{}c@{}}7-shot\\ \end{tabular}} & \multicolumn{1}{c|}{\begin{tabular}[c]{@{}c@{}}10-shot\\ \end{tabular}} \\ \hline

\multicolumn{1}{|c|}{$A=\{R,T_x,T_y\}$}  & \multicolumn{1}{c|}{40.6\% ±4.2}                                                                  & \multicolumn{1}{c|}{	41.4\% ±4.6}                                                               & \multicolumn{1}{c|}{44.7\% ±5.0}                                                                   & \multicolumn{1}{c|}{46.9\% ±4.2}                                                                  &   48.7\% ±3.5                                                                          \\ \hline

\multicolumn{1}{|c|}{$A=\{C,B,S\}$}  & \multicolumn{1}{c|}{47.1\% ±3.5}                                                                  & \multicolumn{1}{c|}{	47.8\% ±5.8}                                                               & \multicolumn{1}{c|}{47.7\% ±5.0}                                                             & \multicolumn{1}{c|}{48.0\% ±5.2}                                                                  &   49.2\% ±4.9                                                                          \\ \hline

\multicolumn{6}{|c|}{Target: EM}                                                                                                                                                  \\ \hline
\multicolumn{1}{|c|}{Augmentation}     & \multicolumn{1}{c|}{\begin{tabular}[c]{@{}c@{}}1-shot\\ \end{tabular}} & \multicolumn{1}{c|}{\begin{tabular}[c]{@{}c@{}}3-shot\\ \end{tabular}} & \multicolumn{1}{c|}{\begin{tabular}[c]{@{}c@{}}5-shot\\ \end{tabular}} & \multicolumn{1}{c|}{\begin{tabular}[c]{@{}c@{}}7-shot\\ \end{tabular}} & \multicolumn{1}{c|}{\begin{tabular}[c]{@{}c@{}}10-shot\\ \end{tabular}} \\ \hline

\multicolumn{1}{|c|}{$A=\{R,T_x,T_y\}$}  & \multicolumn{1}{c|}{47.8\% ±2.6}                                                                  & \multicolumn{1}{c|}{	55.5\% ±1.7}                                                               & \multicolumn{1}{c|}{60.4\% ±3.6}                                                                   & \multicolumn{1}{c|}{64.0\% ±3.2}                                                                  &   67.3\% ±2.7                                                                         \\ \hline

\multicolumn{1}{|c|}{$A=\{C,B,S\}$}  & \multicolumn{1}{c|}{62.0\% ±3.2}                                                                  & \multicolumn{1}{c|}{	69.8\% ±2.7}                                                               & \multicolumn{1}{c|}{70.6\% ±3.3}                                                                   & \multicolumn{1}{c|}{73.1\% ±2.9}                                                                  &   73.7\% ±3.2      \\ \hline
\multicolumn{6}{|c|}{Target: ssTEM}                                                                                                                                                  \\ \hline
\multicolumn{1}{|c|}{Augmentation}     & \multicolumn{1}{c|}{\begin{tabular}[c]{@{}c@{}}1-shot\\ \end{tabular}} & \multicolumn{1}{c|}{\begin{tabular}[c]{@{}c@{}}3-shot\\ \end{tabular}} & \multicolumn{1}{c|}{\begin{tabular}[c]{@{}c@{}}5-shot\\ \end{tabular}} & \multicolumn{1}{c|}{\begin{tabular}[c]{@{}c@{}}7-shot\\ \end{tabular}} & \multicolumn{1}{c|}{\begin{tabular}[c]{@{}c@{}}10-shot\\ \end{tabular}} \\ \hline

\multicolumn{1}{|c|}{$A=\{R,T_x,T_y\}$}  & \multicolumn{1}{c|}{49.4\% ±4.4}                                                                  & \multicolumn{1}{c|}{	61.9\% ±3.4}                                                               & \multicolumn{1}{c|}{64.2\% ±4.0}                                                                   & \multicolumn{1}{c|}{67.6\% ±2.8}                                                                  &   69.7\% ±3.0                                                                                                                                                  \\ \hline

\multicolumn{1}{|c|}{$A=\{C,B,S\}$}  &\multicolumn{1}{c|}{51.3\% ±3.2}                                                                  & \multicolumn{1}{c|}{63.3\% ±3.2}                                                               & \multicolumn{1}{c|}{64.2\% ±3.2}                                                                   & \multicolumn{1}{c|}{67.3\% ±2.9}                                                                  &   68.7\% ±2.6      \\ \hline      
\end{tabular}
\end{center}

\caption{Effect on mIoU results of the data augmentation and selection on the target data set TNBC, EM, and ssTEM.}
\label{tab:aug}
\end{table*}

\begin{table*}[h!]

\begin{center}
\begin{tabular}{|l|c|c|c|c|c|}
\hline
\multicolumn{1}{|c|}{Method}     & \multicolumn{1}{c|}{\begin{tabular}[c]{@{}c@{}}1-shot\\ ($|\mathcal{B}|=100$)\end{tabular}} & \multicolumn{1}{c|}{\begin{tabular}[c]{@{}c@{}}3-shot\\ ($|\mathcal{B}|=300$)\end{tabular}} & \multicolumn{1}{c|}{\begin{tabular}[c]{@{}c@{}}5-shot\\ ($|\mathcal{B}|=500$)\end{tabular}} & \multicolumn{1}{c|}{\begin{tabular}[c]{@{}c@{}}7-shot\\ ($|\mathcal{B}|=700$)\end{tabular}} & \multicolumn{1}{c|}{\begin{tabular}[c]{@{}c@{}}10-shot\\ ($|\mathcal{B}|=1000$)\end{tabular}} \\ \hline
\multicolumn{1}{|c|}{Entropy}  & \multicolumn{1}{c|}{17.3\%±1.3}                                                                            & \multicolumn{1}{c|}{38.5\% ±3.0}                                                                             & \multicolumn{1}{c|}{49.5\% ±4.0}                                                                             & \multicolumn{1}{c|}{54.2\% ±3.7}                                                               &           58.7\% ±3.3                                                                            \\ \hline
\multicolumn{1}{|c|}{MC-dropout} & \multicolumn{1}{c|}{17.2\%±1.3}                                                                             & \multicolumn{1}{c|}{38.8\% ±2.7}                                                                            & \multicolumn{1}{c|}{49.4\% ±3.5}                                                                             & \multicolumn{1}{c|}{54.1\% ±4.0}                                                                             &     58.8\% ±3.1	                                                                                  \\ \hline
\multicolumn{1}{|c|}{Random} & \multicolumn{1}{c|}{20.6\% ±2.0}                                                               & \multicolumn{1}{c|}{39.1\% ±3.7}                                                               & \multicolumn{1}{c|}{	47.2\% ±3.7}                                                               & \multicolumn{1}{c|}{	52.1\% ±4.5}                                                               &      56.1\% ±4.5                                                                                 \\ \hline
\multicolumn{1}{|c|}{\textbf{\textit{Ours}}}  & \multicolumn{1}{c|}{\textbf{25.2\% ±2.0}}                                                       & \multicolumn{1}{c|}{\textbf{41.8\% ±2.9}}                                                     & \multicolumn{1}{c|}{\textbf{52.3\% ±2.4}}                                                     & \multicolumn{1}{c|}{\textbf{55.4\% ±3.0}}                                                     &   \textbf{60.4\% ±2.2}                                                                         \\ \hline \hline
\multicolumn{6}{|c|}{1-shot}                                                                                                                                                  \\ \hline
\multicolumn{1}{|c|}{}                               & \multicolumn{1}{c|}{$|\mathcal{B}|=100$}     & \multicolumn{1}{c|}{$|\mathcal{B}|=200$}   & \multicolumn{1}{c|}{$|\mathcal{B}|=300$}      & \multicolumn{1}{c|}{$|\mathcal{B}|=400$}   &   $|\mathcal{B}|=500$  \\ \hline

\multicolumn{1}{|c|}{Entropy}                               & \multicolumn{1}{c|}{17.3\% ±1.3}     & \multicolumn{1}{c|}{29.9\% ±2.4}   & \multicolumn{1}{c|}{38.4\% ±3.0}      & \multicolumn{1}{c|}{45.9\% ±3.4}   &    48.2\% ±3.2 \\ \hline

\multicolumn{1}{|c|}{MC-dropout}                               & \multicolumn{1}{c|}{17.2\% ±2.0}     & \multicolumn{1}{c|}{29.8\% ±2.4}   & \multicolumn{1}{c|}{38.8\% ±2.7}      & \multicolumn{1}{c|}{45.5\% ±3.2}   &    49.3\% ±3.4 \\ \hline
\multicolumn{1}{|c|}{Random}                               & \multicolumn{1}{c|}{20.6\% ±2.0}     & \multicolumn{1}{c|}{31.1\% ±3.1}   & \multicolumn{1}{c|}{38.5\% ±3.6}      & \multicolumn{1}{c|}{45.1\% ±4.5}   &    47.0\% ±4.8 \\ \hline

\multicolumn{1}{|c|}{\textbf{\textit{Ours}}}                               & \multicolumn{1}{c|}{\textbf{25.2\% ±2.0}}     & \multicolumn{1}{c|}{\textbf{35.1\% ±2.6}}   & \multicolumn{1}{c|}{\textbf{41.8\% ±3.4}}      & \multicolumn{1}{c|}{\textbf{48.8\% ±2.7}}   &    \textbf{51.3\% ±3.2} \\ \hline
\end{tabular}

\end{center}

\caption{Impact of limiting $|\mathcal{B}|$ on the mIoU results for the target data set ssTEM.}
\label{ab:ssTEM_SSS}
\end{table*}

\begin{table*}[t!]

\begin{center}
\begin{tabular}{|lc|c|c|c|c|}
\hline
\multicolumn{1}{|c|}{Method}     & \multicolumn{1}{c|}{\begin{tabular}[c]{@{}c@{}}1-shot\\ ($|\mathcal{B}|=100$)\end{tabular}} & \multicolumn{1}{c|}{\begin{tabular}[c]{@{}c@{}}3-shot\\ ($|\mathcal{B}|=300$)\end{tabular}} & \multicolumn{1}{c|}{\begin{tabular}[c]{@{}c@{}}5-shot\\ ($|\mathcal{B}|=500$)\end{tabular}} & \multicolumn{1}{c|}{\begin{tabular}[c]{@{}c@{}}7-shot\\ ($|\mathcal{B}|=700$)\end{tabular}} & \multicolumn{1}{c|}{\begin{tabular}[c]{@{}c@{}}10-shot\\ ($|\mathcal{B}|=1000$)\end{tabular}} \\ \hline
\multicolumn{1}{|c|}{Entropy}  & \multicolumn{1}{c|}{24.8\%±1.8}                                                                            & \multicolumn{1}{c|}{40.8\% ±0.5}                                                                             & \multicolumn{1}{c|}{45.3\% ±1.3}                                                                             & \multicolumn{1}{c|}{46.7\% ±1.4}                                                               &           48.7\% ±1.3                                                                            \\ \hline
\multicolumn{1}{|c|}{MC-dropout} & \multicolumn{1}{c|}{25.1\% ±1.1}                                                                             & \multicolumn{1}{c|}{41.4\% ±1.4}                                                                            & \multicolumn{1}{c|}{45.3\% ±1.3}                                                                             & \multicolumn{1}{c|}{46.9\% ±1.4}                                                                             &     48.3\% ±1.3	                                                                                  \\ \hline
\multicolumn{1}{|c|}{Random} & \multicolumn{1}{c|}{46.2\% ±2.2}                                                               & \multicolumn{1}{c|}{55.5\% ±3.4}                                                               & \multicolumn{1}{c|}{	61.3\% ±3.5}                                                               & \multicolumn{1}{c|}{	63.3\% ±4.2}                                                               &      64.4\% ±2.3                                                                                 \\ \hline
\multicolumn{1}{|c|}{\textbf{\textit{Ours}}}  & \multicolumn{1}{c|}{\textbf{48.5\% ±1.8}}                                                       & \multicolumn{1}{c|}{\textbf{58.7\% ±3.4}}                                                      & \multicolumn{1}{c|}{\textbf{63.6\% ±2.6}}                                                   & \multicolumn{1}{c|}{\textbf{65.3\% ±2.2}}                                                              &   \textbf{68.4\% ±4.2}                                                                        \\ \hline
\end{tabular}

\end{center}

\caption{Impact of limiting $|\mathcal{B}|$ on the mIoU results for the target data set EM.}
 \label{ab:EM_SSS}
\end{table*}


\begin{table*}[t!]


\begin{center}

\begin{tabular}{|lccccc|}
\hline
\multicolumn{6}{|c|}{Target: TNBC}                                                                                                                                                  \\ \hline
\multicolumn{1}{|c|}{Approach}     & \multicolumn{1}{c|}{\begin{tabular}[c]{@{}c@{}}1-shot\\ \end{tabular}} & \multicolumn{1}{c|}{\begin{tabular}[c]{@{}c@{}}3-shot\\ \end{tabular}} & \multicolumn{1}{c|}{\begin{tabular}[c]{@{}c@{}}5-shot\\ \end{tabular}} & \multicolumn{1}{c|}{\begin{tabular}[c]{@{}c@{}}7-shot\\ \end{tabular}} & \multicolumn{1}{c|}{\begin{tabular}[c]{@{}c@{}}10-shot\\ \end{tabular}} \\ \hline

\multicolumn{1}{|c|}{Expert}  & \multicolumn{1}{c|}{32.4\%}                                                                  & \multicolumn{1}{c|}{41.0\%}                                                               & \multicolumn{1}{c|}{40.0\%}                                                                   & \multicolumn{1}{c|}{41.1\%}                                                                  &   39.6\%                                                                          \\ \hline

\multicolumn{1}{|c|}{\textbf{\textit{Ours}}}  & \multicolumn{1}{c|}{50.3\%}                                                                  & \multicolumn{1}{c|}{50.8\%}                                                               & \multicolumn{1}{c|}{49.0\%}                                                             & \multicolumn{1}{c|}{51.4\%}                                                                  &   53.0\%                                                                        \\ \hline

\multicolumn{6}{|c|}{Target: ssTEM}                                                                                                                                                  \\ \hline
\multicolumn{1}{|c|}{Approach}     & \multicolumn{1}{c|}{\begin{tabular}[c]{@{}c@{}}1-shot\\ \end{tabular}} & \multicolumn{1}{c|}{\begin{tabular}[c]{@{}c@{}}3-shot\\ \end{tabular}} & \multicolumn{1}{c|}{\begin{tabular}[c]{@{}c@{}}5-shot\\ \end{tabular}} & \multicolumn{1}{c|}{\begin{tabular}[c]{@{}c@{}}7-shot\\ \end{tabular}} & \multicolumn{1}{c|}{\begin{tabular}[c]{@{}c@{}}10-shot\\ \end{tabular}} \\ \hline

\multicolumn{1}{|c|}{Expert}  & \multicolumn{1}{c|}{38.0\%}                                                                  & \multicolumn{1}{c|}{55.3\%}                                                               & \multicolumn{1}{c|}{66.0\%}                                                                   & \multicolumn{1}{c|}{68.9\%}                                                                  &   72.4\%                                                                                                                                                 \\ \hline

\multicolumn{1}{|c|}{\textbf{\textit{Ours}}}  &\multicolumn{1}{c|}{52.6\%}                                                                  & \multicolumn{1}{c|}{63.2\%}                                                               & \multicolumn{1}{c|}{60.5\%}                                                                   & \multicolumn{1}{c|}{67.3\%}                                                                  &   70.2\%      \\ \hline      
\end{tabular}
\end{center}

\caption{Effect on mIoU results of the expert selection and our approach on the target data set TNBC and ssTEM. We report the result for only one experiment i.e. we do not average over ten experiments.}
\label{tab:expert}
\end{table*}

\begin{table*}[ht!] 

\begin{center}
    
\begin{tabular}{|lccccc|}
\hline

\multicolumn{6}{|c|}{Target: B39}                                                                                                                                                  \\ \hline

\multicolumn{1}{|c|}{Method}     & \multicolumn{1}{c|}{\begin{tabular}[c]{@{}c@{}}1-shot\\ ($|\mathcal{B}|=100$)\end{tabular}} & \multicolumn{1}{c|}{\begin{tabular}[c]{@{}c@{}}3-shot\\ ($|\mathcal{B}|=300$)\end{tabular}} & \multicolumn{1}{c|}{\begin{tabular}[c]{@{}c@{}}5-shot\\ ($|\mathcal{B}|=500$)\end{tabular}} & \multicolumn{1}{c|}{\begin{tabular}[c]{@{}c@{}}7-shot\\ ($|\mathcal{B}|=700$)\end{tabular}} & \multicolumn{1}{c|}{\begin{tabular}[c]{@{}c@{}}10-shot\\ ($|\mathcal{B}|=1000$)\end{tabular}} \\ \hline
\multicolumn{1}{|c|}{Entropy}  & \multicolumn{1}{c|}{80.9\%±5.6}                                                                    & \multicolumn{1}{c|}{85.9\% ±3.5}                                                                & \multicolumn{1}{c|}{87.3\% ±5.0}                                                                             & \multicolumn{1}{c|}{89.4\% ±2.7}                                                               &           90.1\% ±1.7                                                                            \\ \hline
\multicolumn{1}{|c|}{MC-dropout} & \multicolumn{1}{c|}{72.1\% ±3.0}                                                                             & \multicolumn{1}{c|}{71.3\% ±3.2}                                                                            & \multicolumn{1}{c|}{73.1\% ±3.3}                                                                             & \multicolumn{1}{c|}{76.6\% ±2.6}                                                                             &     77.9\% ±3.9	                                                                                  \\ \hline
\multicolumn{1}{|c|}{Random} & \multicolumn{1}{c|}{\textbf{93.2\% ±0.3}}                                                             & \multicolumn{1}{c|}{\textbf{92.7\% ±0.9}}                                                    & \multicolumn{1}{c|}{	\textbf{92.9\% ±0.5}}                                                               & \multicolumn{1}{c|}{	92.9\% ±0.4}                                                               &      92.8\% ±0.5                                                                                 \\ \hline
\multicolumn{1}{|c|}{\textbf{\textit{Ours}}}  & \multicolumn{1}{c|}{90.1\% ±1.6}                                                                  & \multicolumn{1}{c|}{91.9\% ±1.6}                                                               & \multicolumn{1}{c|}{92.7\% ±0.7}                                                               & \multicolumn{1}{c|}{\textbf{93.0\% ±0.3}}                                                         &   \textbf{92.9\% ±0.4}                                                                          \\ \hline

\multicolumn{6}{|c|}{Target: B5}                                                                                                                                                  \\ \hline
\multicolumn{1}{|c|}{Method}     & \multicolumn{1}{c|}{\begin{tabular}[c]{@{}c@{}}1-shot\\ ($|\mathcal{B}|=100$)\end{tabular}} & \multicolumn{1}{c|}{\begin{tabular}[c]{@{}c@{}}3-shot\\ ($|\mathcal{B}|=300$)\end{tabular}} & \multicolumn{1}{c|}{\begin{tabular}[c]{@{}c@{}}5-shot\\ ($|\mathcal{B}|=500$)\end{tabular}} & \multicolumn{1}{c|}{\begin{tabular}[c]{@{}c@{}}7-shot\\ ($|\mathcal{B}|=700$)\end{tabular}} & \multicolumn{1}{c|}{\begin{tabular}[c]{@{}c@{}}10-shot\\ ($|\mathcal{B}|=1000$)\end{tabular}} \\ \hline
\multicolumn{1}{|c|}{Entropy}  & \multicolumn{1}{c|}{99.1\%±0.1}                                                                    & \multicolumn{1}{c|}{99.1\% ±0.1}                                                                & \multicolumn{1}{c|}{99.1\% ±0.1}                                                                             & \multicolumn{1}{c|}{99.1\% ±0.1}                                                               &           99.0\% ±0.5                                                                            \\ \hline
\multicolumn{1}{|c|}{MC-dropout} & \multicolumn{1}{c|}{99.1\% ±0.1}                                                                             & \multicolumn{1}{c|}{99.1\% ±0.1}                                                                            & \multicolumn{1}{c|}{99.0\% ±0.2}                                                                             & \multicolumn{1}{c|}{99.0\% ±0.1}                                                                             &     99.0\% ±0.2	                                                                                  \\ \hline
\multicolumn{1}{|c|}{Random} & \multicolumn{1}{c|}{\textbf{99.1\% ±0.0}}                                                             & \multicolumn{1}{c|}{\textbf{99.1\% ±0.0}}                                                        & \multicolumn{1}{c|}{\textbf{99.0\% ±0.2}}                                                              & \multicolumn{1}{c|}{\textbf{99.1\% ±0.0}}                                                            &      \textbf{99.1\% ±0.0}                                                                                \\ \hline
\multicolumn{1}{|c|}{\textbf{\textit{Ours}}}  & \multicolumn{1}{c|}{98.9\% ±0.0}                                                                  & \multicolumn{1}{c|}{99.0\% ±0.0}                                                               & \multicolumn{1}{c|}{99.0\% ±0.0}                                                                   & \multicolumn{1}{c|}{99.0\% ±0.0}                                                                  &   99.0\% ±0.0                                                                          \\ \hline

\end{tabular}

\end{center}
\caption{mIoU results for the target testing sets of B39, and B5. Best results are highlighted.}
\label{tab:B39_B5}
\vspace*{6in}

\end{table*}


\end{document}